\documentclass[11pt]{article} 
\usepackage[margin=1in]{geometry} 
\usepackage{setspace}
\usepackage[autostyle, english = american]{csquotes}

\usepackage[normalem]{ulem} 
\usepackage{hyperref}
\usepackage[english]{babel}

\usepackage{amsthm}
\usepackage{amsfonts}
\usepackage{newtxmath}
\usepackage{amsmath}

\newtheorem{proposition}{Proposition}

\usepackage{endnotes}
\let\footnote=\endnote

\usepackage{natbib}
\bibpunct[, ]{(}{)}{,}{a}{}{,}%

\usepackage{color}
\usepackage{url}
\usepackage{pbox}
\usepackage{amsmath}
\usepackage{hyperref}
\usepackage{graphicx}
\usepackage{array}
\newcolumntype{C}[1]{>{\hfil$}p{#1}<{$\hfil}}
\newcommand{\LeftEqNo}{\let\veqno\@@leqno}
\usepackage{environ}
\makeatletter
\NewEnviron{Lalign}{\tagsleft@true\begin{align}\BODY\end{align}}
\makeatother
\usepackage{xcolor}
\usepackage{comment}
\hypersetup{
	colorlinks,
	linkcolor={red!50!black},
	citecolor={blue!50!black},
	urlcolor={blue!80!black}
}
\usepackage{float}
\usepackage[caption = false]{subfig}
\usepackage{graphicx}
\usepackage{tikz}

\usetikzlibrary{arrows}
\usetikzlibrary{calc}
\usepackage{algorithm}
\usepackage[noend]{algpseudocode}
\usepackage{booktabs}
\usepackage{multicol,multirow}
\usepackage{array,longtable}
\usepackage[normalem]{ulem}

\newcommand{\set}[1]{\left\{#1\right\}}

\newcommand{\revision}[1]{\textcolor{black}{#1}}

\newenvironment{Proof}[1][Proof]{\begin{trivlist}
		\item[\hskip \labelsep {\bfseries #1}]}{\end{trivlist}}

\newcommand{\ensemble}{\mathcal{N}}
\newcommand{\nn}{N}
\newcommand{\bnn}{\texttt{B\&C}}
\newcommand{\enn}{\texttt{E-NN}}
\newcommand{\UB}{\texttt{UB}}
\newcommand{\LB}{\texttt{LB}}
\newcommand{\ub}{\texttt{ub}}
\newcommand{\lb}{\texttt{lb}}

\newcommand{\nlayers}{\ensuremath{L}}
\newcommand{\nnodes}[1]{n_{#1}}
\newcommand{\nnns}{e}
\newcommand{\layer}{l}
\newcommand{\neuron}{v}

\newcommand{\y}[3]{y^{#1,#2}_{#3}}
\newcommand{\yvector}[2]{\boldsymbol{y^{#1,#2}}}

\DeclareMathOperator*{\amax}{\arg\!\max}

\usepackage{lscape}
\usepackage{afterpage}
\usepackage{longtable}

\usepackage{setspace}

\allowdisplaybreaks

\begin{document}

\title{Optimizing over an ensemble of neural networks}

\author{Keliang Wang \\ \small Department of Operations and Information Management, School of Business, University of Connecticut
	\and
	Leonardo Lozano  \\ \small Operations, Business Analytics \& Information Systems, University of Cincinnati
	\and
	Carlos Cardonha \\ \small Department of Operations and Information Management, School of Business, University of Connecticut
	\and 
	David Bergman \\ \small Department of Operations and Information Management, School of Business, University of Connecticut
}
\date{}
\maketitle

\abstract{%
	We study optimization problems where the objective function is modeled through feedforward neural networks with rectified linear unit (ReLU) activation. Recent literature has explored the use of a single neural network to model either uncertain or complex elements within an objective function. However, it is well known that ensembles of neural networks produce more stable predictions and have better generalizability than models with single neural networks, which \revision{motivates} the investigation of ensembles of neural networks rather than single neural networks in decision-making pipelines. We study how to incorporate a neural network ensemble as the objective function of an optimization model and explore computational approaches for the ensuing problem. We present a mixed-integer linear program based on existing popular big-$M$ formulations for optimizing over a single neural network. We develop \revision{a two-phase approach for our model that combines preprocessing procedures to tighten bounds for critical neurons in the neural networks with a Lagrangian relaxation-based branch-and-bound approach. Experimental evaluations of our solution methods suggest that using ensembles of neural networks yields more stable and higher quality solutions, compared to single neural networks, and that our optimization algorithm outperforms (the adaption of) a state-of-the-art approach in terms of computational time and optimality gaps.
}%

\maketitle

%

\section{Introduction} \label{sec:introduction}

Finding effective ways of embedding neural networks (NNs) within optimization problems can provide significant improvements in automated decision making.  The potential for application of such a framework \revision{is already explored and discussed both in}  the literature and in practice:
\begin{enumerate}
    \item \emph{Unknown objective}. For example, suppose the revenue for a product is a function of advertising budget and cost, and is modeled through a NN trained from historical data.  Identifying the optimal budget and cost requires embedding the NN in the optimization model. 
    \item \emph{Non-linearity}. If a function does not admit an exact linear representation, a surrogate model for a global optimization problem can be trained from simulated inputs.  This surrogate model, in the form of a NN, can then be used \revision{in a linear model, e.g., } as the objective function.
\end{enumerate}
\revision{Based on the generality of application, finding effective ways of handling NNs within an optimization problem opens the door for enhanced decision making capability.  A major issue is immediately realized: can we trust the output from the NN model at the solution identified by an optimizer? NNs have a tendency to produce highly variable predictions, even for minor changes in inputs, and so optimization problems where the features are variables poses significant risks. }

To address the challenge of highly variable outputs from NNs, we explore in this paper the use of ensemble of NNs in optimization problems instead of single NNs (a preliminary version of this work appears in a recent conference paper~\citep{wang2021two}).  It is well known that ensembles can produce competitive predictive accuracy compared to a single NN with lower variability and better \revision{generalizability}~\citep{Dietterich2000, zhou2002ensembling}, therefore suggesting that their use is particularly well suited for a decision making pipeline.   However, this leads to a challenging optimization problem; optimizing over one NN can be time consuming~\citep{anderson2020}, and existing models \revision{become even more challenging when adapted to solve  ensembles}.

Our focus in this study is therefore to:
\begin{enumerate}
    \item \revision{Introduce the use of NN ensemble to replace single NN within an optimization decision pipeline and demonstrate that NN ensemble can alleviate the issue of variability in predictions; and}
    \item Investigate algorithmic enhancements for optimization models with embedded NN ensembles. 
\end{enumerate}

\revision{To point 1, we evaluate the quality of the enhanced modeling paradigm (i.e., of replacing a single NN by an ensemble of NNs) by comparing the solutions identified.  We show through computations on four surrogate models for global optimization benchmark functions that the solutions obtained through using an ensemble are of higher quality and more stable (i.e., the variance is lower).   
}

\revision{To point 2, we adapt an existing big-$M$ integer programming formulation~\citep{fischetti2018deep, cheng_maximum_2017, anderson2020} and propose a Two Phase algorithm.
The first phase combines a preprocessing procedure that seeks to strengthen the baseline formulation by finding strong bounds for variables associated with a subset of critical nodes and a set of valid inequalities derived from Benders optimality cuts for a decomposition of the problem. In the second phase we develop a Lagrangian relaxation-based branch-and-bound method. We assess the performance of our algorithms using four global optimization benchmark functions and two real-world data sets. The results exhibit a superiority of our algorithm over a benchmark state-of-the-art branch-and-cut approach in terms of computational performance.}

The reminder of the paper is organized as follows. After providing a literature review in Section \ref{sec:literature}, we introduce the notation and a baseline model in Section~\ref{sec:baseline}. Section \ref{sec:enhancements} presents our proposed Two Phase algorithm. We present the results of our experiments in Section \ref{sec:experiments} and conclude with directions for future work in Section \ref{sec:conclusion}.

\section{Literature Review}\label{sec:literature}
We investigate a category of optimization problems where components of the objective function are represented (or approximated) by a Machine Learning model $\hat{f}(x;\theta)$, characterized by a vector~$\theta$ of fixed parameters and a vector~$x$ of input features, whose values can be selected and optimized by the underlying optimization model. \revision{We restrict our attention to ensembles of neural networks with Rectified Linear Unit (ReLU) activation functions, which can be formulated as mixed-integer linear programs (MILP). Other predictive models that have been investigated in 
the optimization literature include logistic regression, linear models, decision trees, random forests, and single neural networks with ReLU activation functions~\citep{bergman2019janos,verwer2017auction,biggs2017optimizing,mivsic2020optimization}}.  \revision{Predictive models can be used to eschew computational intractability by serving as surrogates for complex (e.g., highly nonlinear) functions within an optimization framework \citep{liu_-time_2020, bertsimas2016analytics, xiao_training_2019}}. \revision{Nevertheless, optimizing over a predictive model can be computationally challenging}; for example,  maximizing (or minimizing) the output of random forests or neural networks is NP-hard~\citep{mivsic2020optimization, katz2017reluplex}. 
 
Recent studies on solving optimization models with embedded neural networks have explored different techniques to encode neural networks. \citet{schweidtmann2019deterministic} study neural networks with hyperbolic tangent (\texttt{tanh}) activation functions to solve deterministic global optimization problems; the formulation is solved by a customized branch-and-bound based solver that relies on McCormick relaxations of the~\texttt{tanh} activation functions. \cite{bartolini2011neuron} propose so-called neuron constraints to incorporate NN formulations into a constraint programming approach~\citep{bartolini2011neuron}. Recent work has focused mainly on neural networks with ReLU activation functions, in part because (1) the ReLU function is recommended as a default activation when training neural networks, as it performs well in \revision{numerous} applications~\citep{goodfellow2016deep}; and (2) the piecewise linear nature of ReLU admits a relatively simple
big-$M$ formulation that is easily incorporated within MILP models.

\revision{A recent line of research}
concentrates on using MILPs to verify certain properties of neural networks, \revision{including}
reachability analysis of NNs~\citep{lomuscio_approach_2017}, robustness of NNs to adversarial inputs~\citep{cheng_maximum_2017, tjeng_evaluating_2019, xiao_training_2019, fischetti2018deep}, and output range of a trained NN~\citep{dutta_output_2018}; \revision{the formulations used in these articles} 
are expressed essentially as MILPs (see also \citet{bunel_unified_2018, botoeva_efficient_2020}). \revision{More recently, MILPs have also been used}
for reducing the size of trained neural networks (
\citet{serra2020lossless}) and finding the expressiveness of NNs~\citep{serra2018bounding}.

Furthermore, different techniques have been explored to speed up the solution times for NN-embedded optimization models. As the constant values used in big-$M$ formulations affect the strength of a problem formulation and, consequently, its solution time~\citep{vielma_mixed_2015}, several pre-processing procedures \revision{have been proposed} to identify tighter bounds for the big-$M$ values~\citep{cheng_maximum_2017, dutta_output_2018, tjeng_evaluating_2019,fischetti2018deep, grimstad_relu_2019}; we discuss these methods in Section \ref{sec:enhancements}, as they are closely related to our proposed techniques.\revision{~\cite{lombardi2016lagrangian} apply Lagragian relaxation to a NN with one hidden layer and \texttt{tanh} as activation function, and use the subgradient method to optimize for tighter bounds.}
\cite{bunel_unified_2018} split the input domain to form smaller MILPs restricted to each part of the solution space. \cite{botoeva_efficient_2020} define dependency relations between neurons in terms of their activation (or deactivation) and explore them to derive cuts that reduce the search space. \cite{anderson2020} propose an exponentially-sized convex hull formulation for a single neuron and provide an efficient separation procedure to find the most violated inequality at any given fractional point. \cite{tsay2021} extend the work of \citet{anderson2020} by partitioning the input vector of a ReLU function into groups and considering convex hull formulations over the partitions via disjunctive programming. Depending on the number of partitions, their formulation is able to approximate the convex hull of a ReLU function with fewer number of constraints and auxiliary variables than the \revision{ideal} formulation of~\cite{anderson2020}. 

\revision{Predictive models are typically not exact mappings.
Therefore, there may exist  discrepancies  between the predicted and the actual value of a feasible solution, so predict-and-optimize models may incorrectly prove the optimality of 
sub-optimal solutions (see e.g., \cite{smith2006optimizer}). 
The inaccuracy of predictive models tends to increase in regions of the solution space that are not adequately covered by the samples in the training set, so one \revision{mitigation strategy}
consists of restricting the feasibility  of the optimization problem to well-populated regions of the solution space, i.e., feasible solutions must be close to the existing data points~\citep{biggs2017optimizing,bertsimas2020predictive,maragno_mixed-integer_2021,wasserkrugensuring}.}
\revision{This distance-based condition can be enforced either through the incorporation of constraints~\citep{thebelt2021entmoot,shi2022careful} or as a penalty term in the objective function~\citep{mistry_mixed-integer_2021}. Distance measures used in the area  include  the Euclidean and Manhattan distances~\citep{thebelt2021entmoot}, as well as the Mahalanobis distance and the average distance from the nearest neighbour points~\citep{shi2022careful}. Another strategy consists of restricting the feasible region to the convex hull of the training set (see e.g., \cite{biggs2017optimizing,maragno_mixed-integer_2021}).
}

\revision{Our work focuses on solution methods for ensembles of neural networks.}
\cite{Dietterich2000} provides statistical, computational and representational arguments to show that an ensemble is always able to outperform each of its individual component estimators. Ensembles of neural network were introduced by \citet{hansen1990neural} and have gained substantial development over the last years (see~\citet{li2018research} for a detailed survey). \revision{The idea of exploring NNs as proxies for complex models have been explored in the online learning setting by \cite{lu2017ensemble}; the authors use ensembles 
to develop an adaptation of Thompson sampling to scenarios where the underlying models are intractable. To the best of our knowledge, our work is the first to consider ensembles of NNs for offline optimization. }


\revision{The literature is rich in examples where the integration of predictive models and optimization  has been successfully applied to real-world problems.} \citet{bertsimas2016analytics} optimize over a ridge regression \revision{model} to recommend effective treatments for cancer. \citet{liu_-time_2020} use linear estimators of travel times  to optimize real-time order assignments for a food service provider. Other examples can be found in scholarship allocation for admitted students to maximize class size~\citep{bergman2019janos}, personalized pricing to maximize revenue~\citep{biggs2021model}, and ordering of the items to sell in an auction to maximize expected revenue~\citep{verwer2017auction}. 
\revision{The growing interest in the area has fostered the development of several  software packages that facilitate the development of models based on the integration of optimization and machine learning; some examples are~\cite{EML2017, bergman2019janos,  maragno_mixed-integer_2021}, and \cite{thebelt2021entmoot}.} 

\section{Basic Optimization Model}\label{sec:baseline}

\revision{
The problem we study is as follows. Given a function~$f$, let~$\mathcal{E}_f = \{\nn^1,\nn^2,\ldots,\nn^e\}$ be an ensemble of~$e$ NNs representing~$f$ and $\Omega$ be the feasible set of possible inputs to the ensemble. 
In general, an ensemble is built in two steps: a method to train multiple estimators and the approach to combine their predictions. We adopt the popular Bagging method~\citep{breiman1996bagging} for training NNs and use the averaging method to combine predictions.
An estimate of~$f$ for any point~$x$ in~$\Omega$ produced by~$\mathcal{E}_f$ is given by the average of the individual estimates of each neural network in~$\mathcal{E}_f$, that is,}
\[
\mathcal{E}_f(x) = \frac{1}{e}\sum_{i = 1}^{e}\nn^i(x).
\]
\revision{We study the following problem:}
\begin{equation}\label{ProblemDefinition}
    \max_{x \in \Omega} \mathcal{E}_f(x)
\end{equation}

We now present our baseline optimization model, which was first proposed in~\cite{wang2021two}  and is a straightforward extension of existing MIP models for optimizing over a single NN (see e.g.,~\citet{fischetti2018deep}).   Consider an ensemble~$\ensemble \equiv \{\nn_1,\ldots,\nn_{\nnns}\}$ of~$\nnns$ neural networks. Each NN is a layered graph and we refer to each vertex in a NN as a neuron. Let $\nlayers_i$ denote the number of layers and $\nnodes{\layer}^i$ denote the number of neurons in the $\layer$-th layer of~$\nn_i$, $i \in \set{1,\dots,\nnns}$. We use vector~$(\nnodes{1}^i,\ldots,\nnodes{\nlayers_i}^i)$  to succinctly represent the architecture of~$\nn_i$ and assume that $\nnodes{\nlayers_i}^i = 1$ for all $i \in \set{1,\dots,\nnns}$, i.e., there is only one neuron in the output layer of each NN in~$\ensemble$. We refer to $\{2, 3, \ldots, L_i-1\}$ as the set of intermediate layers of~$\nn_i$, i.e., all layers except the first and the last. Finally, the neural networks of~$\ensemble$  do not share neurons or arcs, so the description of each \revision{neural network}~$\nn_i$ of~$\ensemble$, described next, provides a complete characterization of the ensemble.

Let~$\neuron^{i,\layer}_{j}$ denote the~$j^{th}$ neuron in layer~$\layer$ of~$\nn_i$. Each~$\neuron^{i,\layer}_{j}$   receives a series of inputs and produces a single (scalar) output~$\y{i}{\layer}{j}$.  For the~$j$-th neuron of the first layer, both its input and its output is the~$j$-th coordinate of~$x$, i.e.,  $\neuron^{i,1}_{j}$ receives~$x_j$, the $j^{th}$ component of the decision variables vector, and returns
 \[
 \y{i}{1}{j} = x_j.
 \]
 Observe that all NNs of an ensemble~$\ensemble$ must have the same number of nodes in the first layer, so  we occasionally simplify the notation by dropping the reference to the NN and use~$\nnodes{1}$ instead of~$\nnodes{1}^i$.
 
 Each neuron~$\neuron^{i,\layer}_j$ of an intermediate layer~$\layer$ receives a vector of inputs $\yvector{i}{\layer-1} \equiv \left[\y{i}{\layer-1}{1}, \dots, \y{i}{\layer-1}{\nnodes{\layer-1}^i}\right]$, given by the outputs of the neurons in the previous layer. The output~$\y{i}{\layer}{j}$ of~$\neuron^{i,\layer}_j$  is defined as 
\begin{equation}\label{eq:output_definition}
\y{i}{\layer}{j} = \text{ReLU}\left((\textbf{W}^{i,l}_{j})^{\intercal} \yvector{i}{\layer-1} + b^{i,l}_j\right),
\end{equation}
where~$\text{ReLU}: \mathbb{R} \rightarrow \mathbb{R}^+$ is the Rectified Linear Unit (ReLU) activation function, defined as $ReLU(\bullet) \equiv \max(0,\bullet)$. $\textbf{W}^{i,l}_{j} \in \mathbb{R}^{\nnodes{\layer-1}^i}$ is a \emph{weight} vector and $b^{i,l}_j \in \mathbb{R}$ is a \emph{bias} scalar. Both~$\textbf{W}^{i,l}_{j}$ and~$b^{i,l}_j$ are generated during the training process of~$\nn_i$ and, for the purpose of optimization over~$\ensemble$, they are fixed (i.e., these values cannot be modified).

Finally, the output~$\y{i}{L_i}{1}$ of~$\neuron^{i,L_i}_{1}$, the single terminal neuron in the last layer of~$\nn_i$,  is computed as an affine combination of the previous layer's output without applying the ReLU function, i.e., 
\[
\y{i}{L_i}{1} = (\textbf{W}^{i,L_i}_{1})^{\intercal} \yvector{i}{L_i-1} + b^{i,L_i}_1.
\]

\subsection{Formulation of ReLU functions}

The fundamental building block of our formulation is a mathematical programming representation of the ReLU function for a single neuron, which has been widely used in the literature \citep{fischetti2018deep,tjeng_evaluating_2019,anderson2020}. For each neuron~$\neuron^{i,\layer}_{j}$, we
define an auxiliary continuous variable~$h^{i,\layer}_j$ to capture the linear component of
Expression~\ref{eq:output_definition}, which is the value of the affine combination of the neuron's inputs before applying the ReLU function (also known as the \revision{\emph{pre-activation}}). Additionally, we define a binary variable $z^{i,\layer}_j$ that takes a value of $1$ if~$\neuron^{i,\layer}_{j}$ is \emph{active} (i.e., its output is strictly greater than $0$). Finally, assume that a lower bound~$\LB^{i,\layer}_j < 0$  and upper bound~$\UB^{i,\layer}_j > 0$ are known for each~$h^{i,\layer}_j$, i.e., $h^{i,\layer}_j \in [\LB^{i,\layer}_j,\UB^{i,\layer}_j]$;
observe that neurons for which both bounds are either non-positive or non-negative can be removed or merged~\citep{serra2020lossless}. A MILP formulation that models the behaviour of~$\neuron^{i,\layer}_{j}$ is given by:   
\begin{subequations} \label{SingleNeuron}
\begin{align}
&\quad h^{i,l}_j= (\textbf{W}^{i,l}_j)^{\intercal} \textbf{y}^{i,l-1} + b^{i,l}_j \label{SingleConst1} \\
&\quad h^{i,l}_j \leq y^{i,l}_j\leq h^{i,l}_j - \LB^{i,\layer}_j(1 - z^{i,l}_j) \label{SingleConst2} \\
&\quad 0 \leq y^{i,l}_j \leq \UB^{i,\layer}_j z^{i,l}_j  \label{SingleConst3} \\
&\quad z^{i,l}_j \in \{0, 1\} \\
&\quad h^{i,l}_j,y^{i,l}_j\in \mathbb{R} 
\end{align}
\end{subequations}
Constraint \eqref{SingleConst1} sets the value of~$h^{i,\layer}_j$. 
Constraints \eqref{SingleConst2}--\eqref{SingleConst3} ensure that if $h^{i,\layer}_j > 0$, then $y^{i,\layer}_j = h^{i,\layer}_j$ and $z^{i,\layer}_j = 1$, i.e., they enforce that the output equals the evaluation of the ReLU function.

Observe that $\LB^{i,\layer}_j$ and $\UB^{i,\layer}_j$ act as big-$M$ constants in~\eqref{SingleConst2} and~\eqref{SingleConst3}, respectively.  Therefore,  the strength of Formulation~\eqref{SingleNeuron} for~$\neuron^{i,\layer}_{j}$ strongly depends on the values of 
$\LB^{i,\layer}_j$ and~$\UB^{i,\layer}_j$. One of the main contributions of this work is a strategy to compute tighter bounds, which we present in~\S\ref{sec:strongbounds}.

\subsection{MILP formulation for optimization over ensembles}

We now present an MILP formulation for optimizing over a given ensemble~$\ensemble$, which is a straightforward adaptation of Model \eqref{SingleNeuron} over the complete set of neurons in~$\ensemble$. 
\begin{subequations} \label{BigM}
\begin{align}
\max& \quad\frac{1}{e} \sum_{i = 1}^{e} y^{i,L_i}_1 \label{obj} \\
\text{s.t.}& \quad y^{i, 1}_j  = x_j  &\forall i \in \set{1,\dots,\nnns}, \ j \in \set{1,\dots,\nnodes{1}}  \label{Const1} \\
&\quad y_1^{i,L_i} = h^{i,L_i}_{1}  &\forall i \in \set{1,\dots,\nnns} \label{Const2} \\
&\quad h^{i,l}_j= (\textbf{W}^{i,l}_j)^{\intercal} \textbf{y}^{i,l-1} + b^{i,l}_j &\forall i \in \set{1,\dots,\nnns}, \ l \in \set{2,\dots,\nlayers_i}, \ j \in \set{1,\dots,\nnodes{\layer}^i} \label{Const3} \\
&\quad h^{i,l}_j \leq y^{i,l}_j\leq h^{i,l}_j - \LB^{i,\layer}_j(1 - z^{i,l}_j) &\forall i \in \set{1,\dots,\nnns}, \ l \in \set{2,\dots,\nlayers_i-1}, \ j \in \set{1,\dots,\nnodes{\layer}^i} \label{Const4} \\
&\quad 0 \leq y^{i,l}_j \leq \UB^{i,\layer}_j z^{i,l}_j &\forall i \in \set{1,\dots,\nnns}, \ l \in \set{2,\dots,\nlayers_i-1}, \ j \in \set{1,\dots,\nnodes{\layer}^i} \label{Const5} \\
&\quad z^{i,l}_j \in \{0, 1\} &\forall i \in \set{1,\dots,\nnns}, \ l \in \set{2,\dots,\nlayers_i-1}, \ j \in \set{1,\dots,\nnodes{\layer}^i} \label{Const6} \\
&\quad h^{i,l}_j,y^{i,l}_j\in \mathbb{R} &\forall i \in \set{1,\dots,\nnns}, \ l \in \set{1,\dots,\nlayers_i}, \ j \in \set{1,\dots,\nnodes{\layer}^i} \label{Const7} \\
&\quad x \in \Omega \label{Const8}
\end{align}
\end{subequations}

Variable~$x$ corresponds to the input vector, which belongs to the (potentially constrained) set~$\Omega \subseteq \mathbb{R}^{\nnodes{1}}$. The objective function optimizes the average output of the last neuron in each NN. Constraints \eqref{Const1} ensure that the outputs for neurons in the first layer of any NN correspond to~$x$. Constraints \eqref{Const2} set the output of the single neuron in the last layer of each NN to the affine combination of the neuron's inputs, without applying the ReLU function. Constraints \eqref{Const3}--\eqref{Const7} replicate Formulation~\eqref{SingleNeuron} for each intermediate neuron in the ensemble. 

\subsection{Baseline algorithm}

\citet{anderson2020} investigate Formulation~\eqref{SingleNeuron} and provide \revision{an ideal}   convex hull formulation in the space of the original variables with an exponential number of constraints; they  propose a branch-and-cut algorithm ($\bnn{}$) which iteratively finds and adds the most violated constraints at fractional solutions obtained during the exploration of the branch-and-bound tree. To the best of our knowledge,  \bnn{} is the state-of-the-art approach for optimizing over a single neural network.

As Formulation~\eqref{BigM} is also based on  Formulation~\eqref{SingleNeuron}, \bnn{} 
can be directly adapted to solve our problem. As a result, we use \bnn~as a baseline benchmark algorithm to compare against and measure the effects of our proposed acceleration techniques in~\S\ref{sec:experiments}.       

\section{Two Phase Algorithm}\label{sec:enhancements}

\revision{We propose~\enn{}, a Two Phase algorithm, to optimize over ensemble of neural networks. The first phase relies on acceleration strategies to enhance the formulation of each NN composing the ensemble, whereas the second phase explores a Lagrangian decomposition  of Formulation~\eqref{BigM}. \enn{} is summarized in Algorithm~\ref{twoPhase}.}

\begin{algorithm}[H]
\begin{algorithmic}[1]

\State \revision{\texttt{Pre-Processing:}  Compute $\LB^{i,\layer}_j$ and $\UB^{i,\layer}_j$ using 
Algorithm \ref{TargetedBounds}. }

\State \revision{ \texttt{Phase One:} Execute Formulation \eqref{BigM} enhanced with Inequalities~\eqref{VI} until an optimal solution is found or a time limit is reached. Record the best feasible solution found as $(\bar{x}, \bar{y}, \bar{h}, \bar{z})$. If $(\bar{x}, \bar{y}, \bar{h}, \bar{z})$ is not proven optimal, go to Step 3; otherwise, terminate execution.} 

\State \revision{ \texttt{Phase Two:} Compute the values for Lagrangian multipliers $\lambda$ by applying~$Q$ iterations of a subgradient algorithm. 
Execute the Lagrangian relaxation-based branch-and-bound approach until an optimal solution is found or a pre-defined time limit is reached. }

\end{algorithmic}
\caption{Two Phase Algorithm (\enn{})}
\label{twoPhase}
\end{algorithm}

\subsection{Pre-Processing and Phase One}\label{sec:phase1}

\revision{In the Pre-Processing stage of Algorithm~\ref{twoPhase}, we use the  procedure described in~\ref{sec:strongbounds}  to compute strong bounds for the nodes.
We then try to solve the problem to optimality using Formulation~\eqref{BigM} enhanced with valid inequalities \eqref{VI}, presented in~\ref{sec:valid_ineq}. If we find an optimal solution within a phase-one time limit, then we terminate the algorithm and report an optimal solution. These techniques do not require (or explore) the fact that we have an ensemble, i.e., they handle each NN individually; consequently, the strategies presented in~\ref{sec:phase1} can also be applied to single NNs. }



\subsubsection{Targeted Strong Bounds}\label{sec:strongbounds}

Depending on the tightness of the bounds, the output for intermediate neurons computed at fractional solutions can greatly deviate from the correct evaluation of the ReLU function in Formulation~\eqref{SingleNeuron}. We motivate the need for strong values of $\LB^{i,\layer}_j$ and $\UB^{i,\layer}_j$ with an example.

\smallskip

\noindent \emph{Example 1.} Let~$\neuron^{i,\layer}_j$ be a neuron of an intermediate layer with~$h^{i,\layer}_j \in [-20,10]$. Consider the vector~$(\bar{h}^{i,\layer}_j , \bar{y}^{i,\layer}_j , \bar{z}^{i,\layer}_j)$ of values associated with~$\neuron^{i,\layer}_j$ composing a fractional solution to Model~\eqref{SingleNeuron}. Because the bounds are loose,
the vector $(\bar{h}^{i,\layer}_j , \bar{y}^{i,\layer}_j , \bar{z}^{i,\layer}_j) = (-5, 5, 0.5)$ satisfies constraints \eqref{SingleConst2} and \eqref{SingleConst3}, as 
\begin{align*}
-5 \leq 5 \leq -5 + |-20|(1 - 0.5) \quad \text{and} \quad
0 \leq 5 \leq 10(0.5),
\end{align*}
respectively. Observe that the correct evaluation of the ReLU function applied to  $\bar{h}^{i,\layer}_j=-5$ is  0 (instead of 5). Conversely, if the bounds of~$h^{i,\layer}_j$ were tightened to $[-5,10]$, then the same assignment~$\bar{h}^{i,\layer}_j = -5$ could only compose a feasible solution with
$\bar{y}^{i,\layer}_j = \bar{z}^{i,\layer}_j = 0$, thus yielding  the correct  value of~$\text{ReLU}(\bar{h}^{i,\layer}_j)$. $\blacksquare$ 

\smallskip

A basic procedure for computing bounds from the literature is via interval arithmetic \citep{cheng_maximum_2017, tjeng_evaluating_2019, anderson2020}. In this technique, 
bounds are lexicographically computed layer by layer as:
\begin{align}
\LB^{i,\layer}_j &= \sum_{k \in \set{1,\dots,\nnodes{\layer-1}^i}}\left(\LB^{i,\layer-1}_k\max\{0,w^{i,\layer}_{j,k}\} + \UB^{i,\layer-1}_k\min\{0,w^{i,\layer}_{j,k}\}\right)+b^{i,\layer}_j; \text{ and } \\
\UB^{i,\layer}_j &= \sum_{k \in \set{1,\dots,\nnodes{\layer-1}^i}}\left(\UB^{i,\layer-1}_k\max\{0,w^{i,\layer}_{j,k}\} + \LB^{i,\layer-1}_k\min\{0,w^{i,\layer}_{j,k}\}\right)+b^{i,\layer}_j,
\end{align}
where $w^{i,\layer}_{j,k}$ denotes the scalar value at the $k^{th}$ position of weight vector $\textbf{W}^{i,\layer}_j$ and the bounds for neurons in the first layer are equal to the respective bounds for~$x$.  Interval arithmetic \revision{typically produces} weak bounds, as  over-estimated bounds from one layer are  used in the computation of the bounds for the next layer, thus propagating \revision{errors} 
through the network \citep{tjeng_evaluating_2019, tsay2021}. As an alternative to interval arithmetic, \citet{tjeng_evaluating_2019} propose a  \emph{progressive bounds tightening} procedure that solves up to two linear programs for each neuron~$\neuron^{i,\layer}_j$. These LPs are obtained by relaxing the integrality constraints on the $z$-variables and considering the objectives of maximizing and minimizing $h^{i,\layer}_j$. To obtain even tighter bounds, \cite{fischetti2018deep} solve two MILPs per neuron, obtained by considering the objectives of maximizing and minimizing $h^{i,\layer}_j$ for each critical neuron without relaxing the integrality constraints on the $z$-variables.

The bounding procedures described above offer a trade-off between the computational time required to compute the bounds and the quality of the bounds obtained. Interval arithmetic is quite efficient but provides weak bounds. On the other hand, solving two MILPs per neuron provide strong bounds at the expense of considerably higher computation times. In order to offset the high computational times required to obtain strong bounds, we propose solving two MIPs only for \emph{critical neurons}, which are neurons that are likely to correspond to fractional solutions for which the ReLU function evaluation is vastly overestimated. For the non-critical neurons we compute bounds by solving two LPs as done by~\citet{tjeng_evaluating_2019} and \citet{tsay2021}. 

The cornerstone of our proposed bounding approach is a procedure to identify such critical neurons efficiently.
We start with a version of Formulation~\eqref{BigM} that uses bounds obtained via LPs for all neurons in the neural network. After solving~$K$ nodes of the branch-and-bound tree, we 
identify critical neurons by~\emph{surveying} these nodes. Namely, 
for each fractional solution explored, denoted by $(\bar{x}, \bar{h}, \bar{y}, \bar{z})$, we record the \emph{discrepancy} of each neuron at a given fractional solution as 
\begin{align} \label{discrepancy}
\delta^{i,\layer}_j(\bar{h}, \bar{y})=&
\begin{cases}
\bar{y}^{i,\layer}_j &\text{\quad if } \bar{h}^{i,\layer}_j < 0;   \\
\bar{y}^{i,\layer}_j - \bar{h}^{i,\layer}_j  &\text{\quad if } \bar{h}^{i,\layer}_j \geq 0, 
\end{cases}
\end{align}
which captures the magnitude of the overestimation of the ReLU function by~$(\bar{x}, \bar{h}, \bar{y}, \bar{z})$; observe that the correct evaluation of the ReLU function is 0 in the first case and~$\bar{h}^{i,\layer}_j$ in the second case. \revision{A neuron is critical if its total discrepancy, given by the sum of all discrepancies computed for each of the~$K$ surveyed nodes, is greater than or equal to a given threshold $\tau$.}


After identifying the critical nodes, we identify \revision{their bounds}
by solving the two MILPs described above, which incorporate the integrality constraints on the~$z$-variables and minimize and maximize $h^{i,\layer}_j$, respectively. Observe that, to compute the bounds for a neuron~$\neuron^{i,\layer}_j$, we only need to incorporate the nodes of the first~$\layer-1$ layers (plus~$\neuron^{i,\layer}_j$) in the associated MILP, so the problem is \revision{easier than} the original problem. Nevertheless, solving these sub-problems can be  time-consuming, so we 
use the best bound \revision{obtained within a time limit}  as an over-approximated (yet valid) substitute for the optimal objective value (as done by \citet{fischetti2018deep}). In our computational experiments we find that 
the best bounds from the interrupted MILPs \revision{obtained within 5 seconds} are considerably tighter than the solutions from the LPs. Algorithm \ref{TargetedBounds} summarizes our 
procedure.  
\begin{algorithm}[H]
\begin{algorithmic}[1]

\State Solve two LPs for each neuron to initialize the lower and upper bounds for each neuron.

\State Generate Formulation~\eqref{BigM} using the LP bounds and execute the branch-and-bound algorithm until~$K$ nodes are solved.

\State Survey the $K$ nodes and compute the discrepancies~$\delta^{i,\layer}_j(\bar{h}, \bar{y})$ for each neuron~$\neuron^{i,\layer}_j$ at every fractional solution identified when solving Formulation~\eqref{BigM} with the LP bounds.

\State Identify the set of critical neurons as 
\begin{equation*}
\revision{
\mathcal{C} = \left\{ \neuron^{i,\layer}_j \ | \ \frac{\sum_{k=1}^K \delta^{i,\layer}_j(\bar{h}^k, \bar{y}^k)}{K} \geq \tau, \ i \in \set{1,\dots,\nnns}, \ l \in \set{1,\dots,\nlayers_i}, \ j \in \set{1,\dots,\nnodes{\layer}^i}   \right\}. 
}
\end{equation*}

\State Solve two MILPs for the critical neurons and use best bounds identified within the time limit. 

\end{algorithmic}
\caption{Targeted Strong Bounds Procedure}
\label{TargetedBounds}
\end{algorithm}
After computing the strengthened bounds via Algorithm \ref{TargetedBounds}, we solve Formulation~\eqref{BigM} using the updated bounds. \revision{From this procedure,}
it often happens that several  neurons can be removed,  as \revision{one can infer that} they are always active (when both bounds are non-negative) or they are always inactive (when both bounds are non-positive)~\citep{cheng_maximum_2017,serra2020lossless}.  


\subsubsection{Valid Inequalities}\label{sec:valid_ineq}
We propose a set of valid inequalities for Formulation~\eqref{BigM} that can be interpreted as optimality Bender's cuts 
\citep{Benders62}. Let $\mathcal{Z}$ be the discrete space defined by the integrality constraints \eqref{Const6}, and let $\mathcal{X}(z)$ be the space defined by the remaining constraints \eqref{Const1}--\eqref{Const5} and \eqref{Const7}--\eqref{Const8} for a fixed binary vector $z$. We \revision{rewrite}  Formulation~\eqref{BigM} as 
\begin{align}
v^* = \max_{z \in \mathcal{Z}}  \max_{(x,h,y) \in \mathcal{X}(z)} \quad\frac{1}{e} \sum_{i = 1}^{e} y^{i,L_i}_1, \label{Benders}  \end{align}
where the outer (maximization) problem contains all the binary variables and the inner (minimization) problem is a LP parameterized by the discrete decisions of the outer problem. Because of strong duality, we can \revision{modify}~\eqref{Benders} by replacing the inner maximization problem with its dual minimization problem. Let $\pi^{i,\layer}_j$, $\alpha^{i,\layer}_j$, and $\beta^{i,\layer}_j$ denote the dual variables associated with constraints \eqref{Const3}, \eqref{Const4}, and \eqref{Const5}, respectively. Let $\Psi$ be the dual feasible space (projecting out dual variables with an objective coefficient equal to 0) and note that the dual feasible space is not parameterized by $z$, i.e., the set of feasible dual points remains the same independently of the discrete decisions from the outer problem. By replacing the inner problem by its dual, we can rewrite~\eqref{Benders} as 
\begin{align}
v^* = \max_{z \in \mathcal{Z}}  \min_{(\pi,\alpha,\beta) \in \Psi} \, \sum_{i = 1}^{e}\sum_{\layer = 2}^{\nlayers_i}\sum_{j = 1}^{\nnodes{\layer}^i} b^{i,\layer}_j \pi^{i,\layer}_j + \sum_{i = 1}^{e}\sum_{\layer = 2}^{\nlayers_i-1}\sum_{j = 1}^{\nnodes{\layer}^i} |\LB^{i,\layer}_j|(1-z^{i,\layer}_j)\alpha^{i,\layer}_j + \sum_{i = 1}^{e}\sum_{\layer = 2}^{\nlayers_i-1}\sum_{j = 1}^{\nnodes{\layer}^i} \UB^{i,\layer}_j z^{i,\layer}_j \beta^{i,\layer}_j, \label{BendersDual}  \end{align}
where the dual objective function is parameterized by the discrete decision variables $z$. 
\begin{proposition} \label{p1}
For any feasible dual solution $(\bar{\pi},\bar{\alpha},\bar{\beta}) \in \Psi$,  the following inequality is valid to Formulation~\eqref{BigM}: 
\begin{align} 
\frac{1}{e} \sum_{i = 1}^{e} y^{i,L_i}_1 \leq \sum_{i = 1}^{e}\sum_{\layer = 2}^{\nlayers_i}\sum_{j = 1}^{\nnodes{\layer}^i} b^{i,\layer}_j \bar{\pi}^{i,\layer}_j + \sum_{i = 1}^{e}\sum_{\layer = 2}^{\nlayers_i-1}\sum_{j = 1}^{\nnodes{\layer}^i} |\LB^{i,\layer}_j|(1-z^{i,\layer}_j)\bar{\alpha}^{i,\layer}_j + \sum_{i = 1}^{e}\sum_{\layer = 2}^{\nlayers_i-1}\sum_{j = 1}^{\nnodes{\layer}^i} \UB^{i,\layer}_j z^{i,\layer}_j \bar{\beta}^{i,\layer}_j \label{VI}
\end{align}
\end{proposition}

\begin{Proof}[\textit{Proof}.]
For any $\bar{z} \in \mathcal{Z}$ for which $\mathcal{X}(\bar{z}) \not = \emptyset$  define 
\begin{equation}
v(\bar{z}) = \max_{(x,h,y) \in \mathcal{X}(\bar{z})} \quad\frac{1}{e} \sum_{i = 1}^{e} y^{i,L_i}_1,
\end{equation}
and note that for any feasible primal solution $(\bar{x},\bar{h},\bar{y}) \in \mathcal{X}(\bar{z})$ it holds that
\begin{equation}
\frac{1}{e} \sum_{i = 1}^{e} \bar{y}^{i,L_i}_1 \leq v(\bar{z}).   \end{equation}
Because of strong duality, there exists a dual solution $(\pi^*, \alpha^*, \beta^*) \in \Psi$ for which 
\begin{equation}
v(\bar{z}) = \sum_{i = 1}^{e}\sum_{\layer = 2}^{\nlayers_i}\sum_{j = 1}^{\nnodes{\layer}^i} b^{i,\layer}_j \pi^{*i,\layer}_j + \sum_{i = 1}^{e}\sum_{\layer = 2}^{\nlayers_i-1}\sum_{j = 1}^{\nnodes{\layer}^i} |\LB^{i,\layer}_j|(1-\bar{z}^{i,\layer}_j)\alpha^{*i,\layer}_j + \sum_{i = 1}^{e}\sum_{\layer = 2}^{\nlayers_i-1}\sum_{j = 1}^{\nnodes{\layer}^i} \UB^{i,\layer}_j \bar{z}^{i,\layer}_j \beta^{*i,\layer}_j.     
\end{equation}
Because of weak duality we obtain that 
\begin{equation}
v(\bar{z}) \leq \sum_{i = 1}^{e}\sum_{\layer = 2}^{\nlayers_i}\sum_{j = 1}^{\nnodes{\layer}^i} b^{i,\layer}_j \bar{\pi}^{i,\layer}_j + \sum_{i = 1}^{e}\sum_{\layer = 2}^{\nlayers_i-1}\sum_{j = 1}^{\nnodes{\layer}^i} |\LB^{i,\layer}_j|(1-\bar{z}^{i,\layer}_j)\bar{\alpha}^{i,\layer}_j + \sum_{i = 1}^{e}\sum_{\layer = 2}^{\nlayers_i-1}\sum_{j = 1}^{\nnodes{\layer}^i} \UB^{i,\layer}_j \bar{z}^{i,\layer}_j \bar{\beta}^{i,\layer}_j,     
\end{equation}
for any $(\bar{\pi},\bar{\alpha},\bar{\beta}) \in \Psi$. This concludes the proof. $\blacksquare$
\end{Proof}

There exists an exponential number of valid inequalities \eqref{VI}. Therefore, we propose an iterative approach, where these inequalities are identified and added
every time that a feasible integer solution is explored. \revision{Observe}  that the strength of the valid inequalities is again dependent on the quality of the lower and upper bounds.

\subsection{\revision{Phase Two}}

\revision{If Algorithm~\ref{twoPhase} does not obtain an optimal solution at the end of Phase One, it stores the best feasible solution~$(\bar{x}, \bar{y}, \bar{h}, \bar{z})$ found so far and proceeds to a second phase, in which~\enn{} solves Formulation~\eqref{BigM} using a Lagrangian relaxation-based reformulation. In contrast with the Pre-Processing stage and Phase One, Phase Two applies only to ensembles with two or more NNs.}


\subsubsection{\revision{Lagrangian Relaxation-Based Decomposition}}
\revision{We propose a decomposition approach that is based on the following reformulation of Model~\eqref{BigM}: }
\begin{subequations} \label{varCopies}
\begin{align}
\max& \quad\frac{1}{e} \sum_{i = 1}^{e} y^{i,L_i}_1 \label{objvarCopies} \\
\text{s.t.}& \quad x^{1}_j  = x^i_j  &\forall i \in \set{2,\dots,\nnns}, \ j \in \set{1,\dots,\nnodes{1}}  \label{varCopies1} \\
& \quad y^{i, 1}_j  = x^i_j  &\forall i \in \set{1,\dots,\nnns}, \ j \in \set{1,\dots,\nnodes{1}}  \label{varCopies2} \\
& \quad\eqref{Const2}-\eqref{Const7} \label{varCopies3} \\ 
&\quad x^i \in \Omega \label{varCopies4} &\forall i \in \set{1,\dots,\nnns}.
\end{align}
\end{subequations}
\revision{Model~\eqref{varCopies} includes one copy of the input variables for each NN in the ensemble. Constraints~\eqref{varCopies1} ensure that the~$j$-th input variable to the~$i$-th NN, $i \in \{2,\ldots,\nnns\}$ is equal to~$x^1_{j}$, the $j$-th  input to~$\nn_1$; in words, these constraints force all the NNs of the ensemble to receive the same input; observe that Models~\eqref{BigM} and~\eqref{varCopies} are equivalent.} \revision{We obtain a Lagrangian relaxation of Model~\eqref{varCopies} by moving constraints \eqref{varCopies1} to the objective function. The resulting formulation is  as follows: }
\begin{subequations} \label{lagRelax}
\begin{align}
\max& \quad\frac{1}{e} \sum_{i = 1}^{e} y^{i,L_i}_1 + \sum_{i = 2}^{e} \sum_{j = 1}^{\nnodes{1}} \lambda_{ij}(x^{1}_j - x^i_j)   \label{objlagRelax} \\
\text{s.t.}& \quad  y^{i, 1}_j  = x^i_j  &\forall i \in \set{1,\dots,\nnns}, \ j \in \set{1,\dots,\nnodes{1}}  \label{lagRelax1} \\
& \quad\eqref{Const2}-\eqref{Const7} \label{lagRelax2} \\ 
&\quad x^i \in \Omega \label{lagRelax3} &\forall i \in \set{1,\dots,\nnns}.
\end{align}
\end{subequations}
\revision{Formulation~\eqref{lagRelax} yields a valid upper bound on the optimal objective function of \eqref{BigM} for any given value of the Lagrangian multipliers $\lambda$.} \revision{Formulation~\eqref{lagRelax} is computationally easier to solve than~\eqref{BigM},  as the objective function \eqref{objlagRelax} is the only place in which the NNs of the ensemble interact. On the downside, 
there could be ``disagreements" between the $x$-variables corresponding to different NNs in an optimal solution to~\eqref{lagRelax}, so the solutions identified by~\eqref{lagRelax} associated with each NN may be suboptimal to~\eqref{BigM}. We address this issue by embedding~\eqref{lagRelax} into a branch-and-bound approach.}

\subsubsection{Branching strategy}

\revision{We propose a branching procedure on the input variables $x$, with decisions guided by the degree of disagreement between the copies of the input variables. } \revision{At the root node of our branch-and-bound tree, we solve Formulation~\eqref{lagRelax} allowing all input variables across all NNs of the ensemble to assume any value in its original domains, that is,}
\begin{equation}
x^i_j \in [\lb_j, \ub_j] \quad \forall i \in \set{1,\dots,\nnns}, \ j \in \set{1,\dots,\nnodes{1}}. 
\end{equation}
\revision{The values for $\lb_j$ and $\ub_j$ are derived from the constraints in $\Omega$ if not readily available.}   

\revision{Our proposed branching scheme works as follows. Consider any given node of the branch-and-bound tree in which the $x$-variables take values $\hat{x}^i_j \in [\hat{\lb}_j, \hat{\ub}_j]$, where $\hat{\lb}_j$ and $\hat{\ub}_j$ are the lower and upper bounds for~$x_j$ at the given node, respectively. Before branching, we first check if the local domain of the $x$-variables is already ``small'', i.e., we check if}
\begin{equation}
 \hat{\ub}_j - \hat{\lb}_j \leq \Delta  \quad \forall j \in \set{1,\dots,\nnodes{1}}, \label{domain} 
\end{equation}
\revision{where $\Delta$ is a sufficiently small constant parameter of the algorithm. If \eqref{domain} is true, then we stop branching and complete the exploration of the current node by reverting to the big-M formulation \eqref{BigM} subject to the domain constraints associated with the node, i.e., with~$\Omega$ enriched by}
\begin{equation*}
x_j \in [\hat{\lb}_j, \hat{\ub}_j]  \quad \forall j \in \set{1,\dots,\nnodes{1}}
\end{equation*}

\revision{Otherwise, if the domains of the input variables are still ``large'', we select the branching variable by first computing the  maximum and minimum values of the $\hat{x}$-variables as:}
\begin{equation*}
x^{\texttt{max}}_j = \max_{i \in \set{1,\dots,\nnns}}\{ \hat{x}^i_j \}, \ \forall j \in \set{1,\dots,\nnodes{1}} \text{ and }
x^{\texttt{min}}_j = \min_{i \in \set{1,\dots,\nnns}}\{ \hat{x}^i_j \}, \ \forall j \in \set{1,\dots,\nnodes{1}}
\end{equation*}
\revision{and then selecting the branching variable index~$\hat{\jmath}$ given by }
\begin{equation*}
\hat{ \jmath } = \amax_{ j \in \{1,\dots,\nnodes{1}\} } \{ x^{\texttt{max}}_j - x^{\texttt{min}}_j\}.
\end{equation*}
\revision{Note that $\hat{\jmath}$ is the index for which there is maximum disagreement between the copies of the input variables corresponding to different NNs.} 
\revision{Once $\hat{\jmath}$ is identified, we create two branches. In the left branch we add a new child node with the updated domains given by:}
\begin{equation*}
x_j \in [\hat{\lb}_j, (x^{\texttt{max}}_j + x^{\texttt{min}}_j)/2]  \quad \forall j \in \set{1,\dots,\nnodes{1}},
\end{equation*}
\revision{while in the right branch we add a new child node with the updated domains given by:}
\begin{equation*}
x_j \in [(x^{\texttt{max}}_j + x^{\texttt{min}}_j)/2, \hat{\ub}_j ]  \quad \forall j \in \set{1,\dots,\nnodes{1}}
\end{equation*}

\revision{As it is usually done in branch-and-bound approaches, we define a primal heuristic to obtain feasible solutions (and, in turn, lower bounds on the optimal value) faster. In our approach, we call our primal heuristic every time we solve a branch-and-bound node. Again, let the values of the $x$-variables at a given node be $\hat{x}^i_j$. Our primal heuristic simply runs the original big-M formulation~\eqref{BigM} for~$\nn_1$, the first NN of the ensemble, enriched with the following additional constraints:}
\begin{equation}
x_j \in [\hat{x}^1_j-\epsilon, \hat{x}^1_j+\epsilon]  \quad \forall j \in \set{1,\dots,\nnodes{1}}
\end{equation}
\revision{Value~$\epsilon$ is a small constant given as an input parameter of the algorithm. Observe that our primal heuristic solves the problem only for the first NN, as preliminary experiments suggest that this is computationally more efficient.} \revision{Finally, we explore the branch-and-bound tree by following a standard best-first strategy. Namely, we select the node with the largest upper bound on the objective value as the next node to be explored.}

\subsubsection{\revision{Updating Multipliers and Step Size}}

\revision{Phase Two relies on the identification of Lagrangian multipliers~$\lambda$; we start with $\lambda^0 = 0$ and some step size $\mu^0$. We update the values of the multipliers in each iteration  using a standard subgradient algorithm~\citep{wolsey1999integer}. Namely, at iteration $q \geq 1$, the subgradient algorithm solves~\eqref{lagRelax} and records the optimal solution obtained as~$\bar{x}^q$. The multipliers are then updated as 
\begin{equation}
\lambda^q_{ij} = \lambda^{q-1}_{ij} -\mu^q(\bar{x}^{1q}_j - \bar{x}^{qi}_j)  \quad \forall i \in \set{2,\dots,\nnns}, \ j \in \set{1,\dots,\nnodes{1}}. \label{updateLag}
\end{equation}
Similarly, we update the step size in each iteration as follows:
\begin{equation}
\mu^q = \frac{\mu^{q-1}}{\sqrt{q}}. \label{updateStep}
\end{equation}
We stop the subgradient algorithm after~$Q$ iterations. We then execute our Lagrangian relaxation-based branch-and-bound approach using the $\lambda$-multipliers obtained from the subgradient algorithm, and terminate once an optimal solution is found or a time limit is reached.}

\revision{We remark that the subgradient procedure 
could be time consuming for large problem instances, as it requires solving $Q$ MIPs given by Formulation \eqref{lagRelax}. To accelerate these computations, we fix the values of the binary variables to be $z = \bar{z}$, where~$\bar{z}$ is part of the best feasible solution obtained in Phase One. With this, we can solve Formulation~\eqref{lagRelax} as an LP, which allows to obtain reasonable initial values for the $\lambda$-multipliers quickly. Moreover, since formulation \eqref{lagRelax} is solved at every node of the branch-and-bound tree, we propose to keep updating the $\lambda$-multipliers and the step size according to \eqref{updateLag} and \eqref{updateStep} as we execute the branch-and-bound approach. Note that by doing this we (potentially) refine the quality of the $\lambda$-multipliers at the beginning of the exploration and since the step size value is reduced with every update, the change to the multipliers becomes negligible as the search explores deeper nodes in the branch-and-bound tree.}


\section{Computational Results}\label{sec:experiments}
We present in this section the results of our computational experiments. In section \ref{subsec:instances} we describe the test problems used in our computations. \revision{In section~\ref{subsec:solqualityCompare}, we compare the solution quality between single NN and NN ensemble using four benchmark functions with known optimal values. In section~\ref{sec:comp_algs}, we compare our approach with a state-of-the-art algorithm on the four benchmark functions mentioned above and two additional real-world datasets. Section \ref{sec:Sensitivity} presents sensitivity analysis and additional experiments to assess the components of our two-phase approach. }

We use Python 3.8,  \texttt{Tensorflow}~(\cite{tensorflow2015-whitepaper}), and a 2.6 GHz 6-Core Intel i7-9750H CPU with 32 GB of RAM to train the neural networks 
of our data sets. 
We use Java implementations of~\bnn{} and~\enn{}, and we use Gurobi 9.0.2 to solve the mixed-integer programming formulations~\citep{gurobi}. The experiments are executed on an Intel Xeon E5--1650 CPU (six cores) running at 3.60 GHz with 32 GB of RAM on Windows 10. Each execution is restricted by a time limit of $3,600$ seconds. All code and instances are  available upon request.  

\subsection{Test Problems}\label{subsec:instances}


\revision{We use six data sets in our experiments. Four benchmark problems were extracted from the global optimization literature,
and the other two are extracted from real-world applications that have been used in the Machine Learning and Optimization communities.
}  

\medskip

\paragraph{\texttt{Peaks}:} The peaks function~$f(x_1,x_2)$ is a benchmark instance in the global optimization literature (\cite{schweidtmann2019deterministic}). Peaks is defined over~$\left[-3,3\right]^2$ as
\begin{equation} \label{peaks}
    p(x_1,x_2) \equiv 3(1 - x_1^2)^2e^{-x_1^2 - (x_2+1)^2} - 10\left(\frac{x_1}{5} - x_1^3 - x_2^5\right)e^{-x_1^2-x_2^2} - \frac{e^{-(x_1+1)^2-x_2^2}}{3}.
\end{equation}
The goal is to identify a solution~$(x_1^*,x_2^*)$ such that~$p(x_1^*,x_2^*)$ is minimum. The global optimal solution value  is~$p(\textbf{x}^*)  =  -6.551$, which is attained at~$x_1^* = 0.228$ and $x_2^* = -1.626$. 

\paragraph{\texttt{Beale}:} \revision{The Beale function is a continuous, non-differentiable, and multimodal function~(\cite{jamil2013literature}) defined over~$[-4.5,4.5]^2$ and given by}
\begin{equation} \label{beale}
    b(x_1, x_2) \equiv (1.5 - x_1+ x_1x_2)^2 + (2.25 - x_1 + x_1x_2^2)^2 + (2.625 - x_1 + x_1x_2^3)^2
\end{equation}
\revision{
The global minimum solution is~$b(\textbf{x}^*) = 0$, which is attained at~$x^*_1 = 3, x^*_2 = 0.5$. }

\paragraph{\texttt{Perm}:} \revision{The Perm function is a parameterized function (\cite{mishra2006global}). We consider the 3-dimensional version defined over~$[-3,4]^3$ as}
\begin{equation}
    m(x_1,x_2,x_3) \equiv \sum_{k=1}^3 \left\{ \sum_{j=1}^3 \left(j^k
        + \frac{1}{2}\right) \left[ \left(\frac{x_j}{j}\right)^k - 1 \right] \right\}^2
\end{equation}
\revision{The global minimum is~$m(x^*)=0$, which is attained at~$x^*=(1,2,3)$.}

\paragraph{\texttt{Spring}:} \revision{The Deflected Corrugated Spring function is another parameterized function (\cite{mishra2006global}). We consider the 5-dimensional instance defined over~$[0,8]^4$ and given by}
\begin{equation} \label{spring}
    s(x_1,x_2,x_3,x_4,x_5) \equiv 0.1\sum_{i=1}^5(x_i - 4)^2 - \cos \left\{ 4\sqrt{\sum_{i=1}^5(x_i - 4)^2} \right\}
\end{equation}
\revision{
The global minimum solution is $s(x^*) = -1$, which is attained at $x^*=(4,4,4,4,4)$.}


\paragraph{\texttt{Wine}:} This first real-world application is based on the wine preference data set introduced by~\cite{cortez2009modeling} (see also~\cite{mivsic2020optimization}). In this work, the authors propose regression techniques to estimate the quality of a wine based on 11 \revision{features}, such as the concentration of residual sugar and the relative volume of alcohol, using a data set with 1,599 samples. In the optimization version of the problem, we wish to identify the physicochemical properties of a wine that would have the highest quality.

\paragraph{\texttt{Concrete}:} This second real-world application was introduced by~\cite{yeh1998modeling}, who proposed the use of neural networks to predict the compressive strength of high-performance concrete based on 8 features, such as the densities of cement and water in the mixture, using 1,030 samples (see also~\cite{mivsic2020optimization}). The optimization problem associated with \texttt{Concrete} can be interpreted as the identification of a concrete composition with maximum compressive strength. 

\medskip

Differently from \revision{\texttt{Beale}, \texttt{Spring}, \texttt{Perm}, and} \texttt{Peaks}, optimal solutions for \texttt{Wine} and \texttt{Concrete} are unknown. Moreover, there is no simple \revision{(i.e., purely analytical)} way of evaluating the quality  of the solutions produced by our algorithms \revision{for these problems}, as closed-formula expressions to evaluate the objective values  are not available. 

\subsection{\revision{ Comparison of Solution Quality}}\label{subsec:solqualityCompare}
\revision{ We experiment on synthetic datasets sampled from the instances for which we know optimal solutions:  \texttt{Peaks}, \texttt{Beale}, \texttt{Perm}, and \texttt{Spring}. Specifically, we train a neural network on the sampled dataset to approximate the underlying function and then solve the optimization problem \eqref{ProblemDefinition}.
The advantage of using benchmark functions for these experiments is two-fold: 
\begin{enumerate}
    \item We can calculate the true value of the optimal solution using a closed-form formula,  and we also know the true global optimizer; and
    \item Some benchmark functions resemble real-world problems with respect to the difficulty of training a good predictive model and finding the true global optimizer.
\end{enumerate} 
In the following section, we discuss how we create test instances for NNs/NN ensembles for optimization, and then provide a detailed comparison of the quality of the solutions using both methods.}

\subsubsection{\revision{Training and Optimization of Neural Networks}}
\paragraph{Generation of  datasets.} \revision{We generate~$2000 + 1000 \times (n-2)$ data points and calculate their respective objective values for each benchmark problem and each data set used for testing, where~$n$ is the number of dimensions. Thus, there are 2000, 2000, 3000, and 5000 data points for~\texttt{Peaks}, ~\texttt{Beale},~\texttt{Perm}, and~\texttt{Spring}, respectively. 
We adopt two sampling strategies to generate data sets; (a) \emph{Latin Hypercube Sampling} (LHS) and (b) generating from a multivariate normal distribution, with mean given by the optimal solution and covariance matrix randomly drawn from the collection of symmetric positive-definite matrices (using \texttt{make\_spd\_matrix()} in the \texttt{sklearn} package). There are thus two data sets for each benchmark problem and therefore a total of eight data sets.  For each data set we employ preprocessing to scale both the inputs and the output into range $[0,1]$.}

\paragraph{Model Selection} 
\revision{\texttt{Tensorflow}~\citep{tensorflow2015-whitepaper} is used to train the neural networks and a combination of \texttt{Tensorflow} and \texttt{BaggingRegressor}~\citep{pedregosa2011scikit} is used to train the NN ensembles. We utilize \texttt{RandomizedSearchCV}~\citep{pedregosa2011scikit} to conduct a hyper-parameter search of neural network models; specifically, we tune the number of hidden layers~$L$, the number of neurons $n$ in hidden layers, learning rate, the batch size, and the number of NNs~$e$ composing the  ensemble; we use~\texttt{Adam} optimizer for training and default settings in~\texttt{Tensorflow} for other hyper parameters. 
For every data set, we conduct a  search for a single NN and another one for an ensemble. Each search randomly samples 1000 combinations of hyper parameters and computes validation scores using $K$-fold cross validation. We then pick the four best combinations based on validation score. In the following sections, we let $(e, L, n)$ represent a specific selected combination.
}
\paragraph{Training and Optimization of Instances}
\revision{For each configuration~$(e, L, n)$, we randomly select 80\% of the data set for training and use the remaining 20\% as the test set. During training, we set aside 20\% of the training set for Early Stopping to prevent over-fitting. We also log the Root Mean Squared Error (RMSE) on the test set. We finally solve the problem~\eqref{ProblemDefinition} with the trained model embedded and evaluate the quality of the solutions. The above training and optimization process is repeated to create 20 replications. Therefore, there are a total of 1280 experiment runs:  4 (four benchmark problems)~$\times$~2 (two sampling strategies)~$\times$~2 (singe neural network and ensemble)~$\times$~4 (four hyper parameter combinations)~$\times$~20 (replicas). We set a $3,600$ seconds time limit for optimization; for instances that cannot be solved to optimality within time limit, we use the best feasible solution found returned by the solver.}

\subsubsection{\revision{Analysis of Solution Quality }}
\revision{For instances trained on LHS sampled data sets, we evaluate the quality of the optimal solutions by calculating the actual function value of the optimal solutions and comparing it against the true global minimum. The results are shown in Figure~\ref{fig:actual_value_distribution}, where each plot represents the results of a benchmark problem. On the horizontal axis, we report the configuration~$(e, L, n)$ atop the corresponding average test RMSE of 20 replications. The vertical axis represents the actual function value. Each boxplot shows the distribution of actual values of the 20 replicas for each configuration, with blue representing ensembles and orange representing single NNs. The red-dotted horizontal line is the true global minimum value. We also add a green interval to indicate the average actual values and its 95\% confidence interval constructed by Bootstrapping.}
\begin{figure}
    \centering
    \includegraphics[scale=0.47]{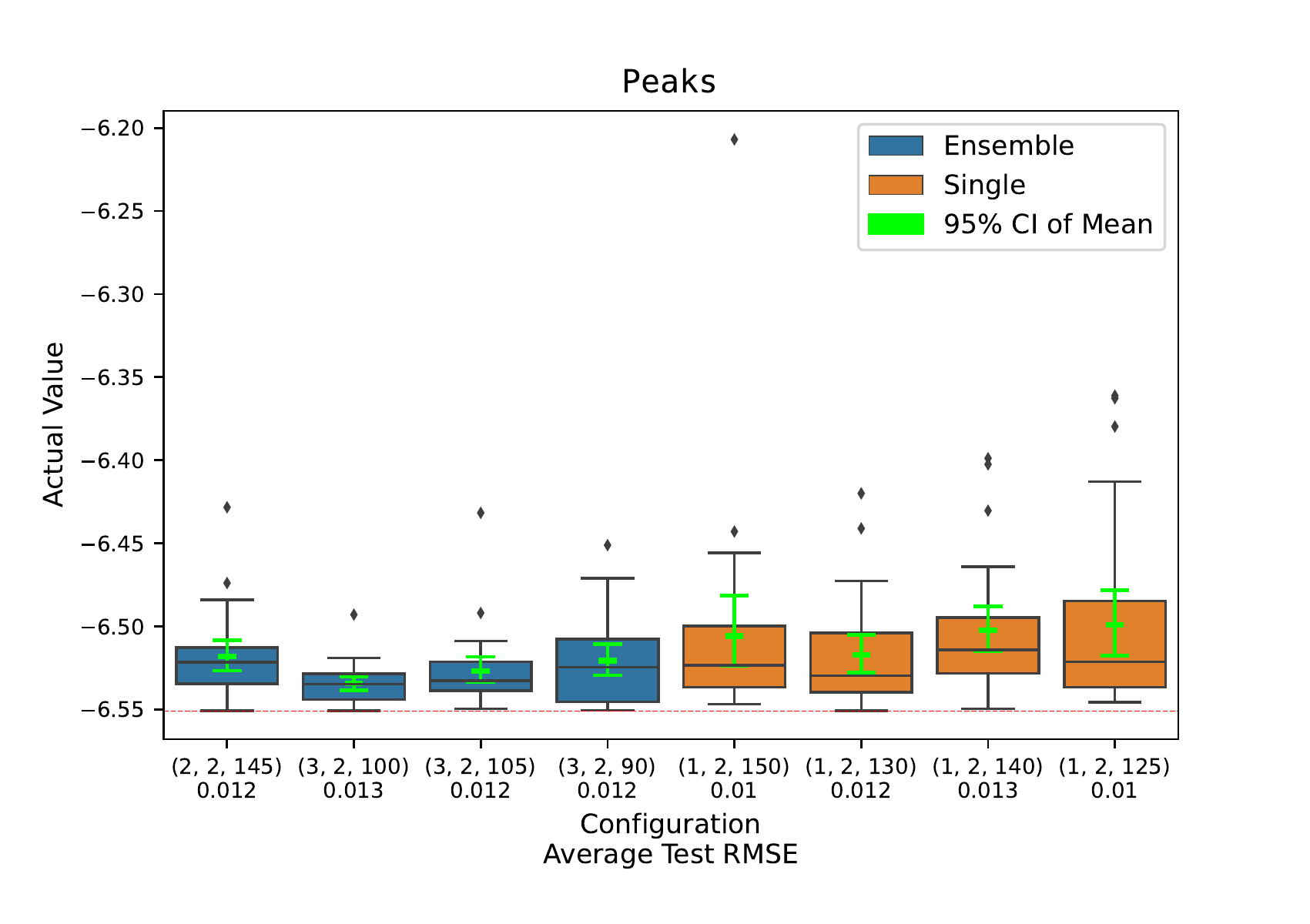}
    \hfill
    \includegraphics[scale=0.47]{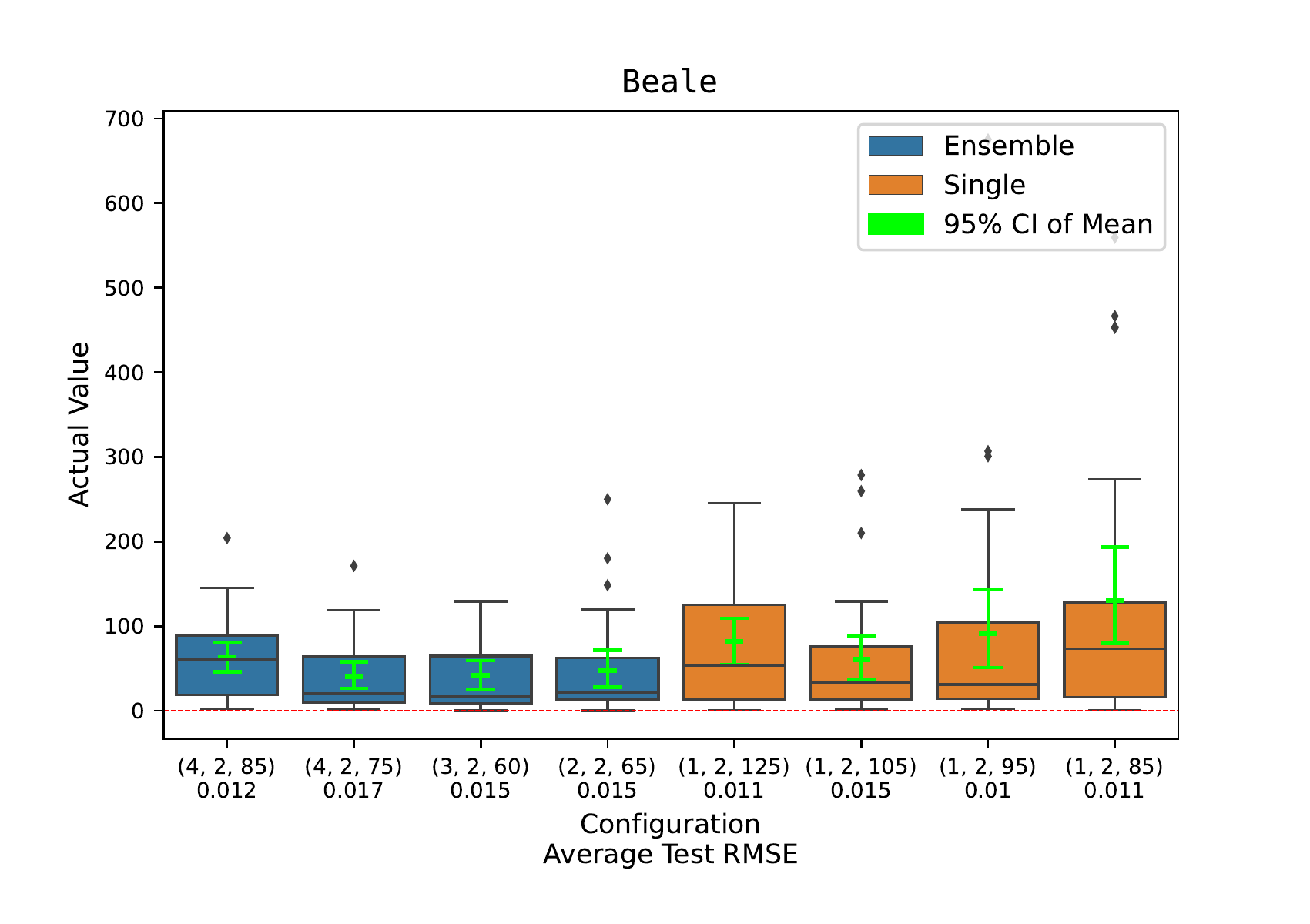}
    \hfill
    \includegraphics[scale=0.47]{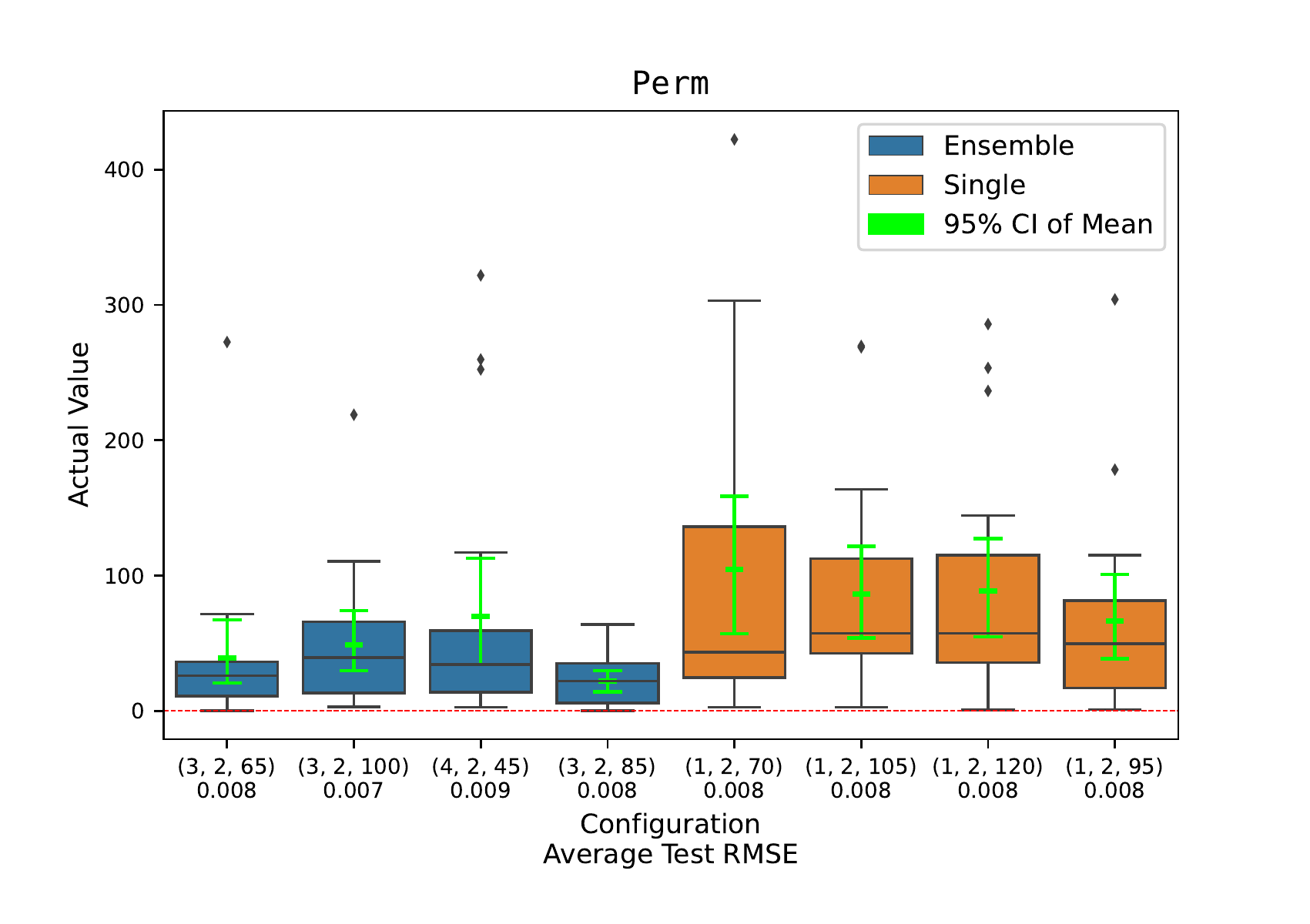}
    \hfill
    \includegraphics[scale=0.47]{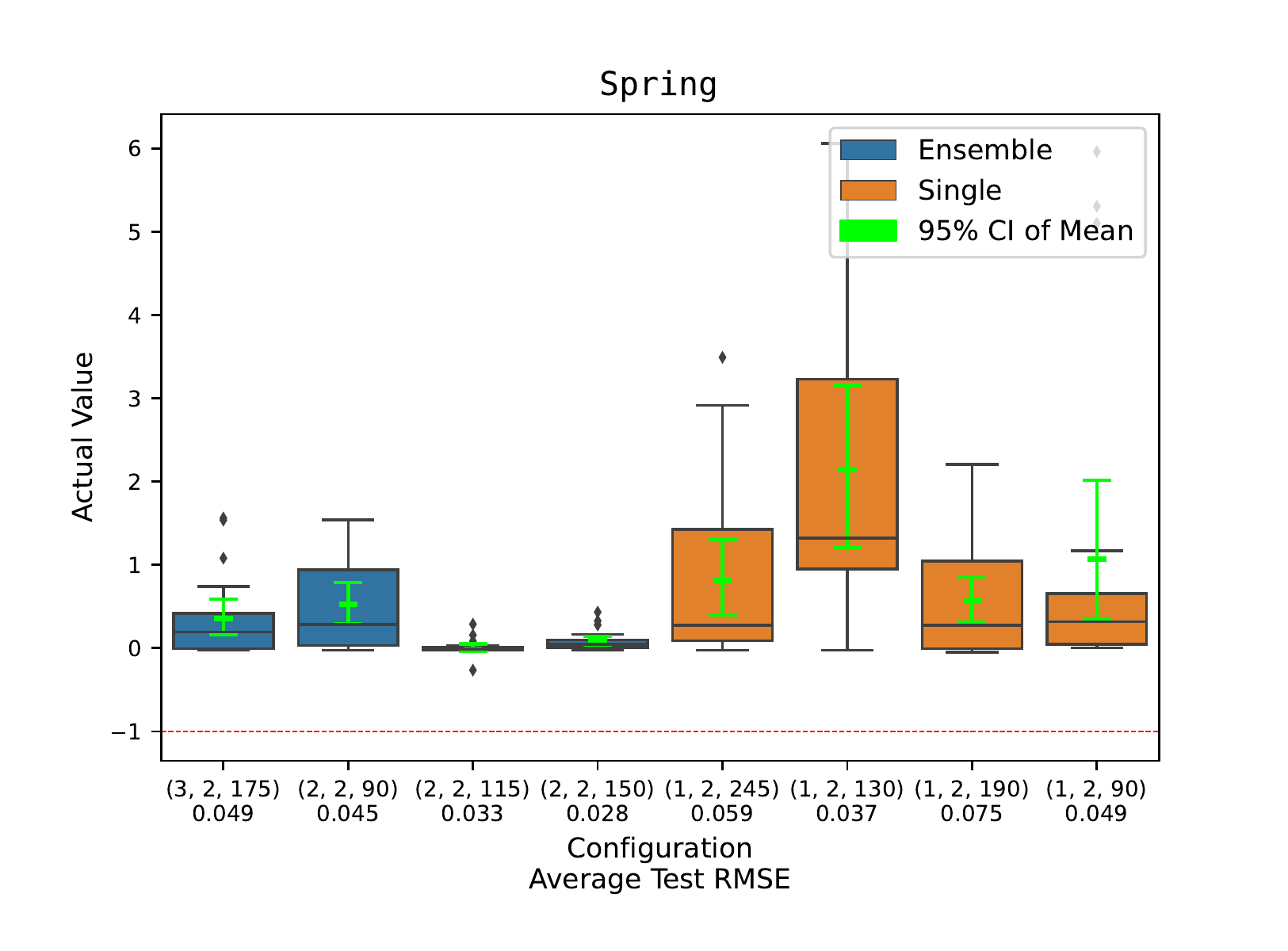}
    \caption{Distribution of Actual Values of Optimal Solutions and 95\%  Confidence Interval of the Mean}
    \label{fig:actual_value_distribution}
\end{figure}

\revision{One significant observation from Figure~\ref{fig:actual_value_distribution} is that the sizes, i.e., inter-quantile range (IQR), of the boxes associated with ensemble models are almost always smaller than that of single neural networks' across the benchmark problems, which suggests that the ensemble-based models is more stable.
The lower variability resulting from using ensembles can also be observed by comparing the 95\% confidence interval (CI) of the mean. Another observation is that the median and mean of the actual values obtained from optimizing with ensembles are closer to the true global minimum than when using single neural networks, i.e., the ensemble-based models generate higher quality solutions.
The average test RMSE between ensembles and single NNs are very close, thus ensuring a fair comparison. 
}


\begin{figure}[H]
    \centering
    \includegraphics[scale=0.47]{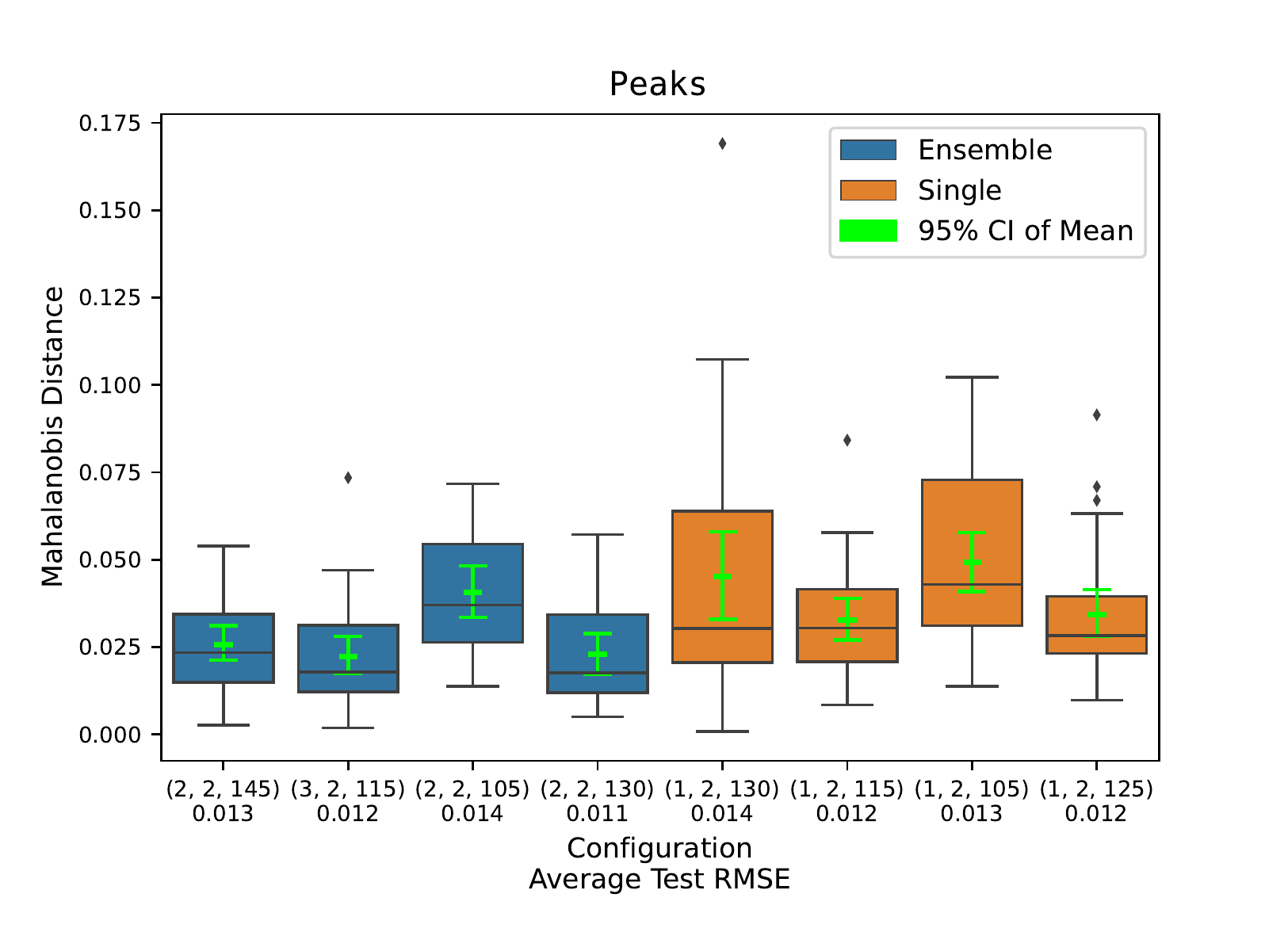}
    \hfill
    \includegraphics[scale=0.47]{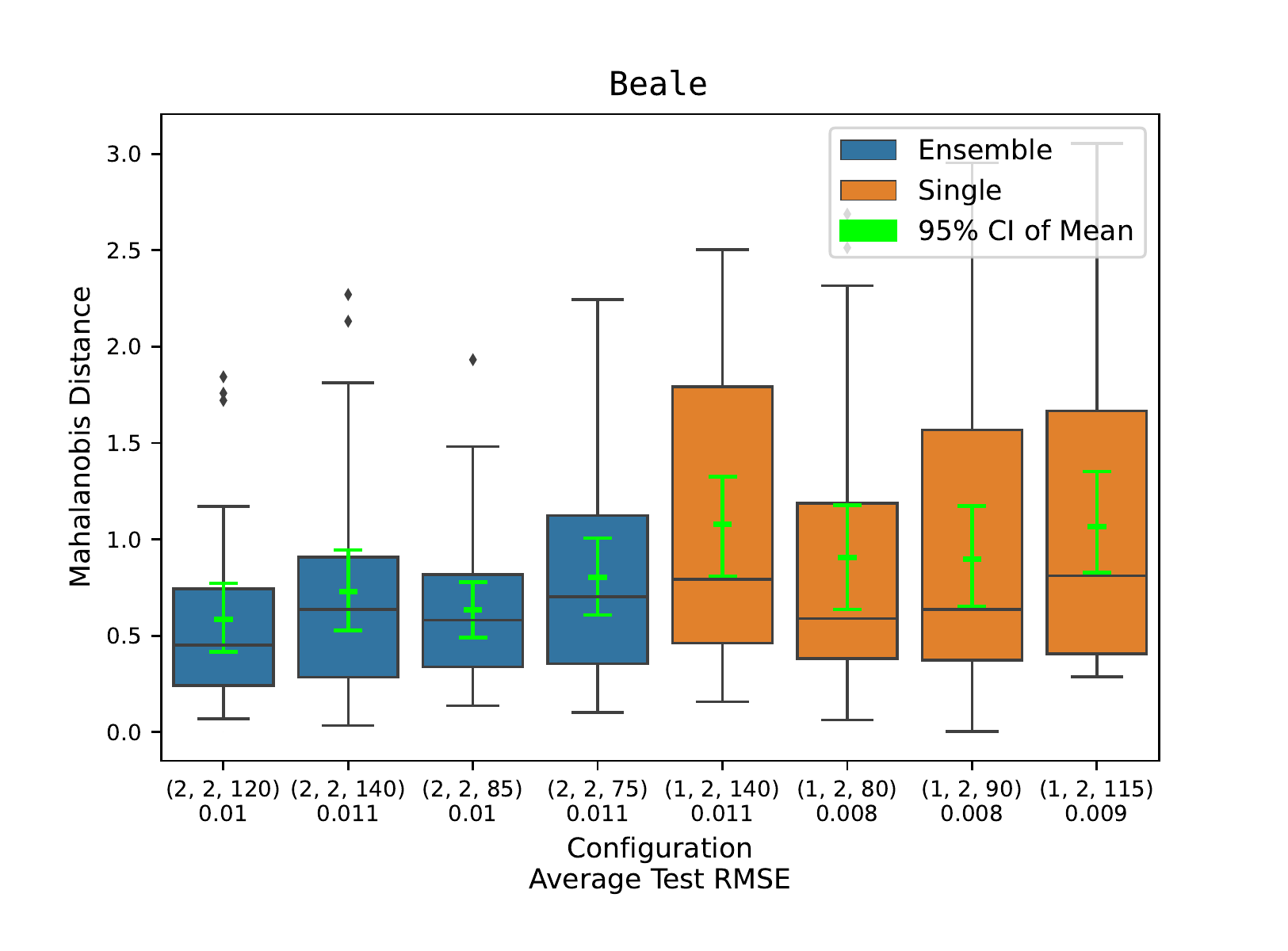}
    \hfill
    \includegraphics[scale=0.47]{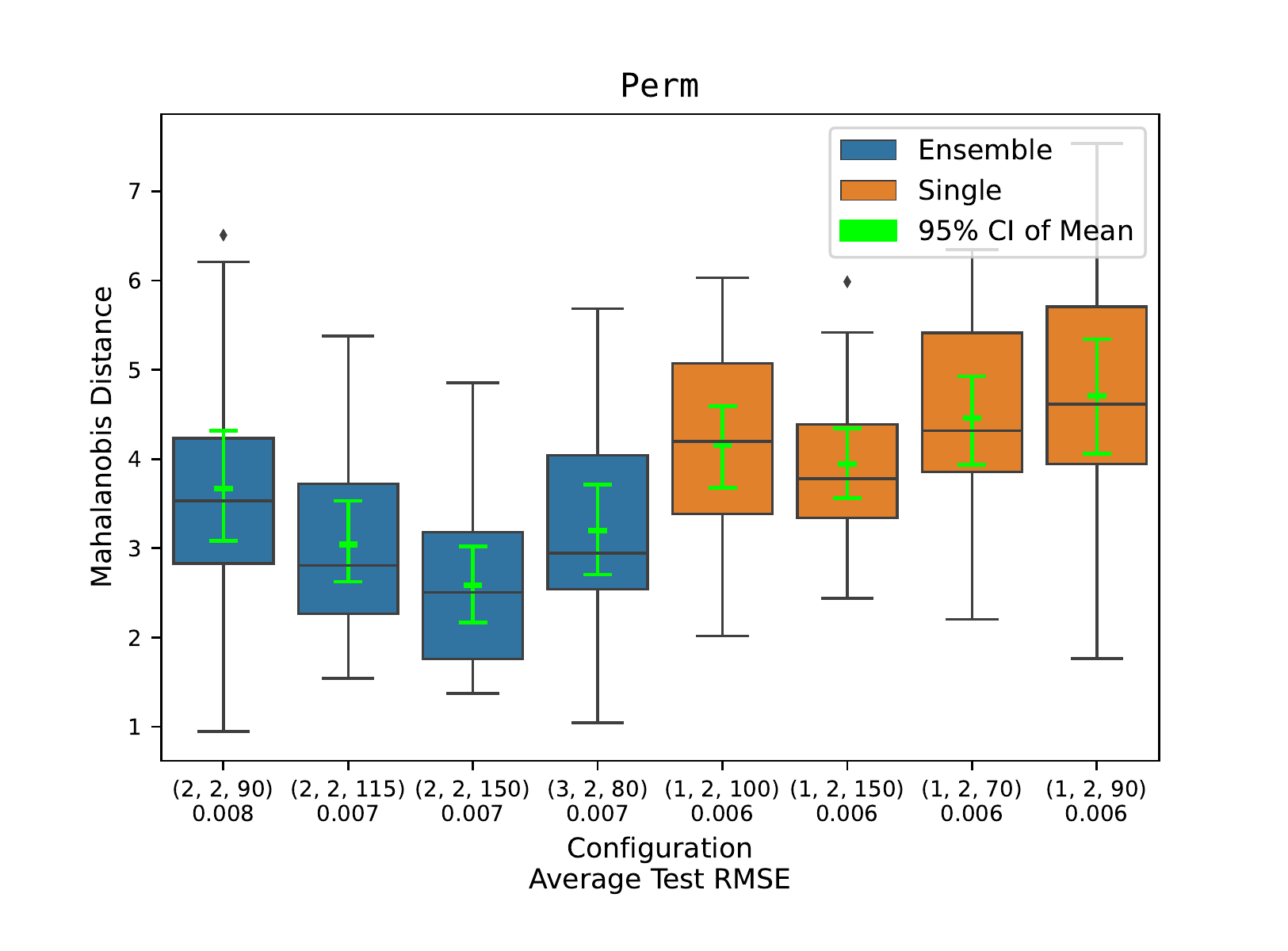}
    \hfill
    \includegraphics[scale=0.47]{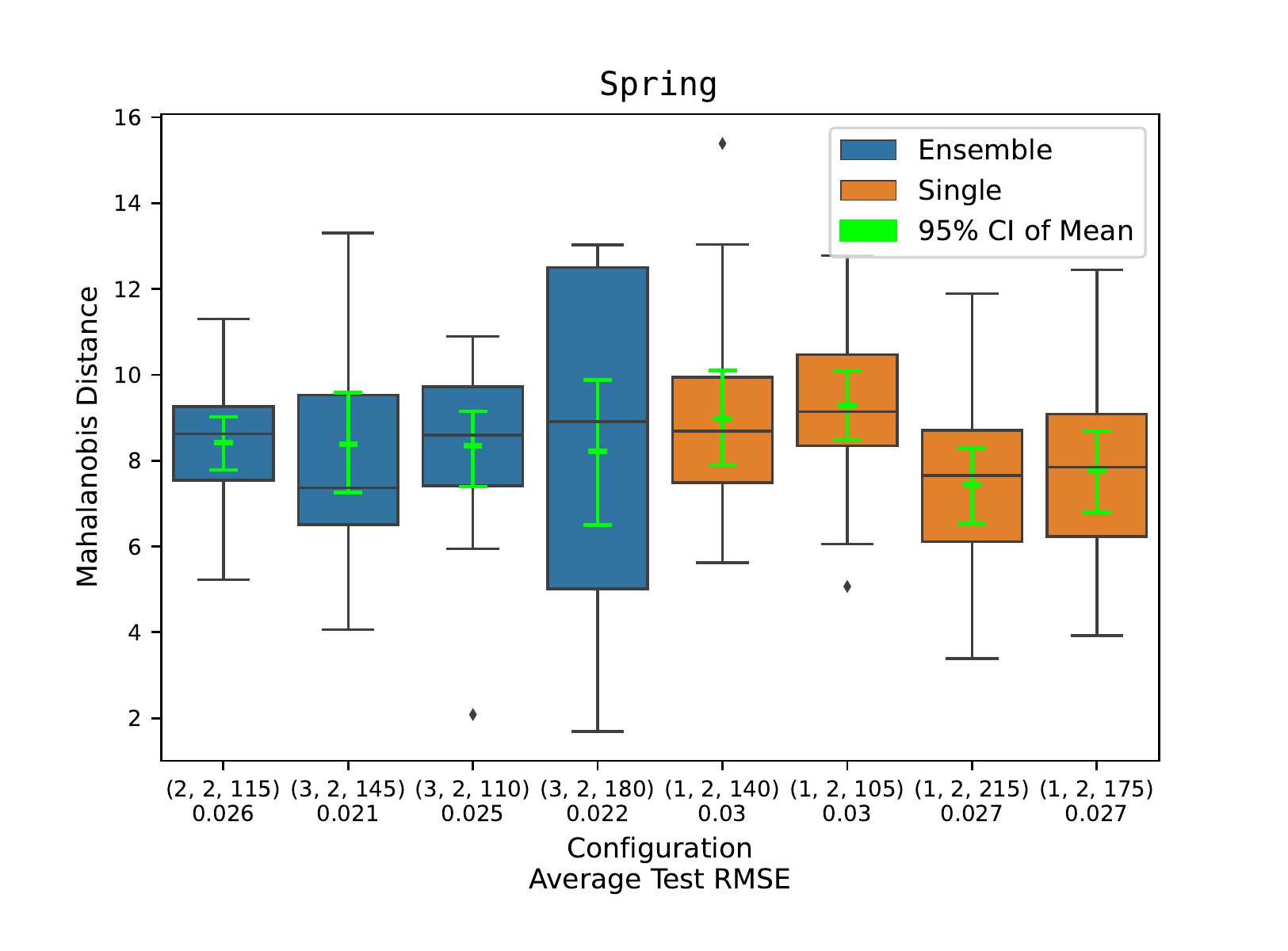}
    \caption{Distribution of Mahalanobis Distance and 95\% Confidence Interval of the Mean}
    \label{fig:M_distance_comparison}
\end{figure}

\revision{Of critical importance when using trained models with optimization frameworks is training relevance, i.e., the solutions obtained should be similar and ideally from the same distribution of the solutions used to train the models. For instances trained on data sets sampled from a multivariate normal distribution, we use the  distance between the optimal solution and the points in the data set to measure the training relevance~\citep{thebelt2021entmoot, mistry_mixed-integer_2021}; specifically, we use the \emph{Mahalanobis distance}.
The results are displayed in Figure~\ref{fig:M_distance_comparison} in boxplots similar to those in Figure~\ref{fig:actual_value_distribution}, but with the vertical axis now representing the Mahalanobis distance. For~\texttt{Peaks}, three out of four ensembles have lower median Mahalanobis distances than every single NNs' median value, and their 95\%  CI lie fully below the lower end of the first and third single NN's 95\%  CIs. For~\texttt{Beale}, ensembles have smaller IQR and average Mahalanobis distance than single NNs. For~\texttt{Perm}, all ensembles have a lower median than single NNs.  For~\texttt{Spring}, we see a mostly comparable collection of solutions, but one ensemble leads to a wide confidence interval.   The above observation indicates that ensemble-based models generate solutions closer to historical data, and thus more reliable. Observe that we did not incorporate distance constraints in our methods to enforce these results. }

\revision{We note here that for \texttt{Peaks}, \texttt{Beale}, and \texttt{Perm} we are reporting results identified by an implementation of model~\ref{BigM}.  For~\texttt{Spring}, however, we are depicting the solutions obtained by our two-phase optimization model since it is a considerably harder problem.}

\subsection{Comparison with the State-of-the-Art Algorithm}\label{sec:comp_algs}
\subsubsection{Test Instances}

We generate three random problem instances for each combination of~$\nnns \in \{3,5\}$, ~$\nlayers \in \{2,4\}$ and $\nnodes{} \in \{20,40\}$, and for each of the \revision{six} benchmark problems described above \revision{resulting in \revision{144} instances}. Observe that \revision{for~$(5,4,40)$, the largest configuration, there are $800$ neurons in the hidden layers} controlled by a ReLU activation function, which in turns correspond to the number of auxiliary binary variables in the formulation.

\subsubsection{Implementation details}

We implement the state-of-the-art branch-and-cut approach ~\bnn{} using callbacks, as described in \cite{anderson2020}. In preliminary computational experiments we found that imposing an upper limit on the total number of generated cuts helps to reduce the total computation time. We set this upper limit to $25,000$ and compute the values for $\LB$ and $\UB$ by solving 2 LPs for each neuron.

Two-phase algorithm $\enn$ uses our proposed targeted strong bounds procedure to compute the values of $\LB$ and $\UB$, and generates valid inequalities \eqref{VI} via lazy cuts every time that a feasible integer solution is found. After fine tuning the parameters for $\enn$ on a subset of instances, we set $K = 1000$, $\tau = 0.01$, and a time limit of $5$ seconds for the MILPs in the targeted strong bounding procedure. \revision{For the first phase we set a time limit of $180$ seconds. For the second phase, we set the initial step size $\mu^0 = 0.05$, $Q=20$, $\Delta = \epsilon = 0.02$. We set a time limit of $3,600$ seconds for the entire optimization process (which includes the time spent in the first phase). The computational times reported in figures and tables include any time spent in pre-processing, computing bounds, separating cuts or valid inequalities, and finding $\lambda$-multipliers.   }

\subsubsection{Performance of Optimization Algorithm}

The performance of~\bnn{} and~\enn{}  is summarized in Figures~\ref{fig:cumulative_plot} and~\ref{fig:scatter_plot} and Table~\ref{tab:results}. Figure~\ref{fig:cumulative_plot} shows a cumulative plot for~\bnn{} and~\enn{} in terms of execution time and optimality gap. Namely, on the left side of Figure~\ref{fig:cumulative_plot}, each point in a curve indicates the number of instances ($y$-axis) that were solved to optimality by the respective algorithm within the amount of time indicated in the~$x$-axis. On the right side, the curves  show the number of instances solved within the optimality gap indicated in the~$x$-axis. 
\revision{We use $\log$ scale on both sides of the cumulative plot.}
\begin{figure}[ht]
\centering
\captionsetup{justification=centering}
\includegraphics[width=0.65\textwidth]{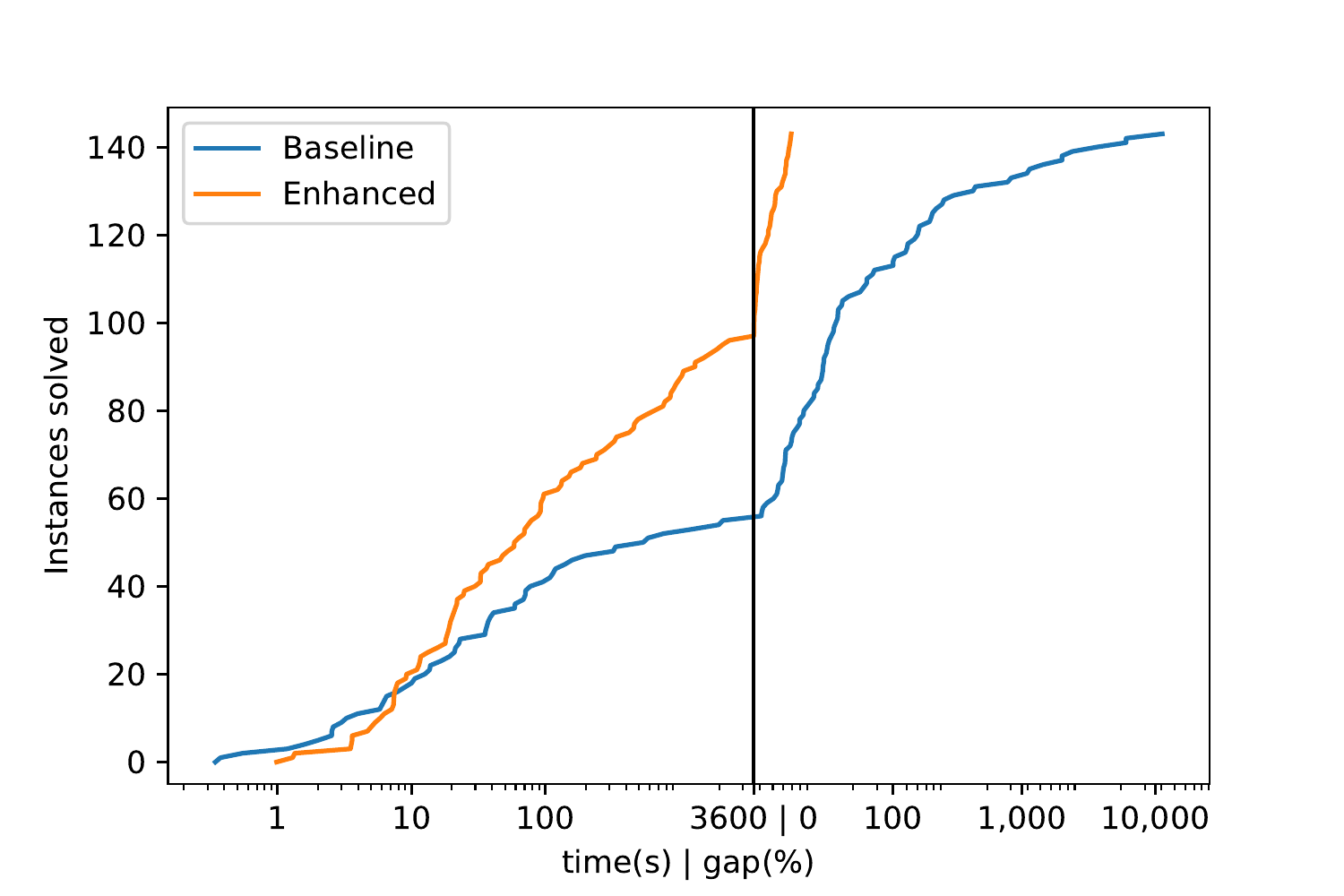}
\caption{Cumulative performance plot comparing~\bnn{} and~\enn{} on all instances.}
  \label{fig:cumulative_plot}
\end{figure}

Figure~\ref{fig:cumulative_plot} shows that the cumulative performance of~\enn{} is superior to that of~\bnn{}. Whereas both algorithms solve a significant number of instances within seconds, \revision{with a minor advantage of~\bnn{} on the easier ones, the performance of~\bnn{} clearly degrades on harder instances.} These observations can be further investigated  with the support of Table~\ref{tab:results}, which aggregates  the results by problem, number~$\nnns$ of ensembles,  and number~$\nlayers$ of intermediate layers; these parameters are presented in the first three columns of the table. The table omits the number~$\nnodes{}$ of nodes per layer, and as we have three random instances per configuration, each entry aggregates the results of 6 instances. For each algorithm we report the average execution times (in seconds, with fractional values rounded up, and recording 3600 seconds for instances that reached the time limit), the number of instances solved to optimality, and the average gap of the instances that were not closed within 3,600 seconds.

 \begin{table}[ht]
\centering
\scriptsize
\caption{Consolidated results}
\label{tab:results}
\begin{tabular}{ccc|ccc|ccc}
\toprule
\multicolumn{3}{c|}{\textbf{Instance}}  & \multicolumn{3}{c|}{\textbf{Baseline (\bnn{})}}  & \multicolumn{3}{c}{\textbf{Enhanced (\enn{})}} 
\\
Instance & $\nnns$ & $\nlayers$  & \textbf{Time} & \textbf{Solved} & \textbf{Gap} & \textbf{Time} & \textbf{Solved} & \textbf{Gap} 
\\
\midrule
\texttt{Peaks} & 3 & 2 &      22 &         6 &      0 &      13 &         6 &      0 \\
     &   & 4 &    1980 &         3 &    269 &     432 &         6 &      0 \\
     & 5 & 2 &      98 &         6 &      0 &      34 &         6 &      0 \\
     &   & 4 &    3600 &         0 &    293 &     755 &         6 &      0 \\
\texttt{Beale} & 3 & 2 &       6 &         6 &      0 &       4 &         6 &      0 \\
     &   & 4 &    1814 &         3 &   4242 &      80 &         6 &      0 \\
     & 5 & 2 &      16 &         6 &      0 &      12 &         6 &      0 \\
     &   & 4 &    1889 &         3 &  27563 &     202 &         6 &      0 \\
\texttt{Perm} & 3 & 2 &    1024 &         5 &     41 &      36 &         6 &      0 \\
     &   & 4 &    3353 &         1 &    177 &    1639 &         4 &      4 \\
     & 5 & 2 &    1831 &         3 &     68 &     490 &         6 &      0 \\
     &   & 4 &    3600 &         0 &    185 &    2028 &         3 &      4 \\
\texttt{Spring} & 3 & 2 &    1410 &         4 &     20 &      38 &         6 &      0 \\
     &   & 4 &    3097 &         1 &    948 &    1886 &         3 &     41 \\
     & 5 & 2 &    1820 &         3 &    116 &     110 &         6 &      0 \\
     &   & 4 &    3600 &         0 &   3283 &    2046 &         3 &     44 \\
\texttt{Concrete} & 3 & 2 &    1820 &         3 &    166 &    1068 &         5 &      5 \\
     &   & 4 &    3600 &         0 &   1022 &    3600 &         0 &     46 \\
     & 5 & 2 &    3600 &         0 &    418 &    3006 &         2 &      2 \\
     &   & 4 &    3600 &         0 &   1078 &    3600 &         0 &     48 \\     
\texttt{Wine} & 3 & 2 &    1839 &         3 &    247 &    1954 &         3 &     10 \\
     &   & 4 &    3600 &         0 &  26354 &    3600 &         0 &     44 \\
     & 5 & 2 &    3600 &         0 &   1242 &    2655 &         2 &     29 \\
     &   & 4 &    3600 &         0 &  21655 &    3600 &         0 &     62 \\
\bottomrule
\end{tabular}
\end{table}

The results presented in Table~\ref{tab:results} show that our enhanced algorithm scales better than the baseline. \revision{Namely, \enn{} solves more instances to optimality in all data sets and} delivers considerably better gaps for instances with larger values of~$\nlayers$ and~$\nnns$. \revision{Observe that the}  relative performances of the algorithms are essentially equivalent \revision{across the different types of problem. For example, \texttt{Peaks} is one of the easiest problems; however, \bnn{} could not solve any instances with 5 neural networks and 4 layers while \enn{} solves all the instances within the time limit. The real-world problems are considerably harder; \texttt{Wine} is the most challenging problem, which can be explained by the number of input features.  Whereas~\enn{} does deliver relatively high optimality gaps for some of the hardest instances, the plots in Figure~\ref{fig:cumulative_plot} show that~\bnn{} can be orders of magnitude worse.}
\begin{figure}[ht]
\centering
\captionsetup{justification=centering}
\includegraphics[width=0.65\textwidth]{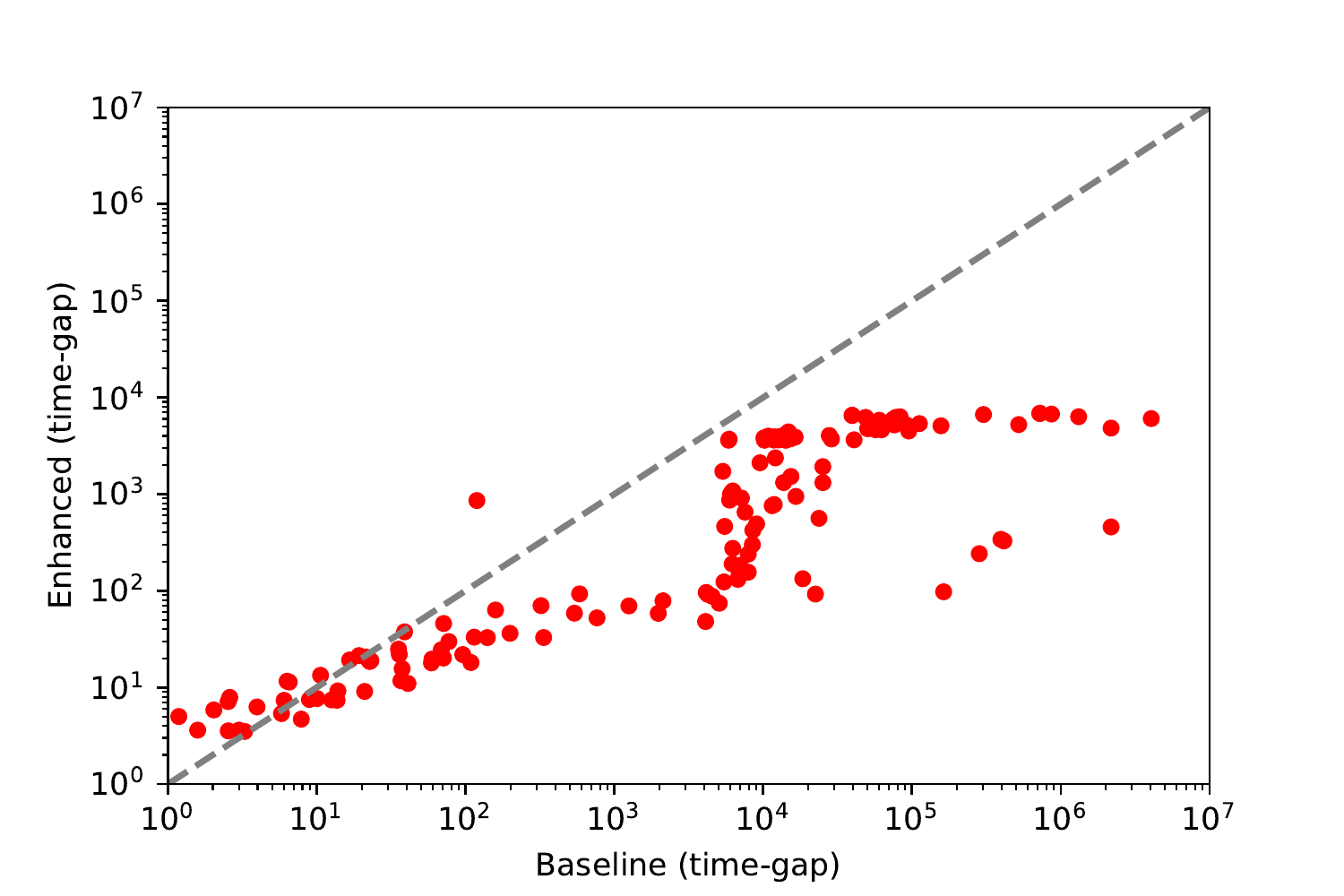}
\caption{Scatter plot comparing execution time-gaps of~\bnn{} and~\enn{} on all instances.}
  \label{fig:scatter_plot}
\end{figure}

\revision{Figure~\ref{fig:scatter_plot} is a scatter plot that compares the performance of~\bnn{} and~\enn{}. To account for differences in performance both in terms of time and gap, we present the results using \textit{time-gap} as the metric describing the performance of each algorithm and on each instance. More precisely, the time-gap metric is given by the expression~$t + 3,600*\alpha$, where~$t$ is the time spent by the algorithm to solve the instance and~$\alpha$ is the optimality gap achieved after 3,600 seconds. Observe that the time-gap metric reduces to the time~$t$ if the instance is solved within the time limit, and that~$t$ is capped at 3,600 otherwise. As in Figure~\ref{fig:cumulative_plot}, the results are presented in $\log$ scale. Although the results of the plot are affected by the multiplicative factor 3,600 applied to~$\alpha$, the differences in performance between the algorithms are clear. \bnn{} is superior for some of the easier instances (those that could be solved within 5 minutes), but the rightmost part of the plot shows that~\enn{} largely outperforms~\bnn{} in harder instances (by orders of magnitude in some cases), for which the algorithms could not solve the problem to optimality.  }

\revision{We now analyze the execution time for the different stages of both algorithms for Beale, Spring, and Wine (complete results are presented in the Appendix~\ref{app:Results}). Table \ref{tab:BCresultsBreakDown} 
presents the results for~\bnn{}.}
\revision{We aggregate the three replicas associated with each configuration, given by the dataset, number of neural networks, number of layers, and number of nodes, which are presented in the first 4 columns. The last five columns report the results. The first indicates the number of replicas  that were solved to optimality by~\bnn{}, whereas the other four columns report the average solution time (in seconds, capped at 3,600 for each experiment), optimality gap (as reported by Gurobi), pre-processing time, and time spent with the branch-and-cut algorithm, respectively.}
 \begin{table}[ht]
\centering
\scriptsize
\caption{Computational Time Break Down for the Baseline algorithm}
\label{tab:BCresultsBreakDown}
\begin{tabular}{llll|rrrrr}
\toprule
     &   &   &    &  Solved &  Time &    Gap (\%) &  Preprocess &  B\&B Time \\
Instance &  $\nnns$ & $\nlayers$ & Nodes &         &       &        &             &           \\
\midrule
\texttt{Beale} & 3 & 2 & 20 &       3 &     1 &      0 &           0 &         0 \\
     &   &   & 40 &       3 &     8 &      0 &           1 &         8 \\
     &   & 4 & 20 &       3 &    29 &      0 &           1 &        27 \\
     &   &   & 40 &       0 &  3600 &   4242 &           7 &      3593 \\
     & 5 & 2 & 20 &       3 &     5 &      0 &           0 &         5 \\
     &   &   & 40 &       3 &    27 &      0 &           2 &        25 \\
     &   & 4 & 20 &       3 &   178 &      0 &           3 &       175 \\
     &   &   & 40 &       0 &  3600 &  27563 &          19 &      3581 \\
\texttt{Spring} & 3 & 2 & 20 &       3 &     2 &      0 &           0 &         2 \\
     &   &   & 40 &       1 &  2817 &     20 &           2 &      2816 \\
     &   & 4 & 20 &       1 &  2595 &    106 &           2 &      2592 \\
     &   &   & 40 &       0 &  3600 &   1508 &          20 &      3580 \\
     & 5 & 2 & 20 &       3 &    40 &      0 &           1 &        39 \\
     &   &   & 40 &       0 &  3600 &    116 &           4 &      3596 \\
     &   & 4 & 20 &       0 &  3600 &    282 &           7 &      3593 \\
     &   &   & 40 &       0 &  3600 &   6284 &          53 &      3547 \\
\texttt{Wine} & 3 & 2 & 20 &       3 &    78 &      0 &           0 &        78 \\
     &   &   & 40 &       0 &  3600 &    247 &           2 &      3598 \\
     &   & 4 & 20 &       0 &  3600 &    266 &           3 &      3597 \\
     &   &   & 40 &       0 &  3600 &  52442 &          25 &      3575 \\
     & 5 & 2 & 20 &       0 &  3600 &    103 &           1 &      3599 \\
     &   &   & 40 &       0 &  3600 &   2382 &           6 &      3594 \\
     &   & 4 & 20 &       0 &  3600 &  21991 &           8 &      3592 \\
     &   &   & 40 &       0 &  3600 &  21320 &          45 &      3555 \\
\bottomrule
\end{tabular}
\end{table}
\revision{We note that the pre-processing time for~\bnn{} is a considerably small fraction of the total execution time across all instance configurations. All problems with 3 neural networks, 2 layers, and 20 nodes are solved in under 80 seconds; however, increasing the number of layers or the number of nodes has a dramatic effect on the performance of the algorithm. Instances not solved to optimality within the time limit exhibit exorbitant optimality gaps.}    

\revision{Table~\ref{tab:2phaseresultsBreakDown} shows the results of~\enn{}. The first eight columns are similar to their counterparts in  Table~\ref{tab:BCresultsBreakDown}. Columns ``Phase 1'' and ``Phase 2'' indicate the average time spent by~\enn{} in Phases 1 and 2, respectively. The last column shows the number of nodes explored in Phase 2.}
 \begin{table}[ht]
\centering
\scriptsize
\caption{Computational Time Break Down for the Enhanced algorithm.}
\label{tab:2phaseresultsBreakDown}
\begin{tabular}{llll|rrrrrrr}
\toprule
     &   &   &    &  Solved &  Time &  Gap (\%)  &  Preprocess &  Phase 1 &  Phase 2 &  BB Nodes \\
Instance & $\nnns$ & $\nlayers$ & Nodes &         &       &      &             &          &          &           \\
\midrule
\texttt{Beale} & 3 & 2 & 20 &       3 &     1 &    0 &           1 &        0 &        0 &         0 \\
     &   &   & 40 &       3 &     7 &    0 &           6 &        1 &        0 &         0 \\
     &   & 4 & 20 &       3 &    16 &    0 &          12 &        5 &        0 &         0 \\
     &   &   & 40 &       3 &   144 &    0 &         106 &       38 &        0 &         0 \\
     & 5 & 2 & 20 &       3 &     5 &    0 &           4 &        0 &        0 &         0 \\
     &   &   & 40 &       3 &    20 &    0 &          16 &        4 &        0 &         0 \\
     &   & 4 & 20 &       3 &    29 &    0 &          24 &        4 &        0 &         0 \\
     &   &   & 40 &       3 &   375 &    0 &         312 &       64 &        0 &         0 \\
\texttt{Spring} & 3 & 2 & 20 &       3 &     7 &    0 &           6 &        1 &        0 &         0 \\
     &   &   & 40 &       3 &    69 &    0 &          44 &       25 &        0 &         0 \\
     &   & 4 & 20 &       3 &   172 &    0 &         144 &       28 &        0 &         0 \\
     &   &   & 40 &       0 &  3600 &   41 &        1198 &      180 &     2284 &        12 \\
     & 5 & 2 & 20 &       3 &    18 &    0 &          13 &        5 &        0 &         0 \\
     &   &   & 40 &       3 &   202 &    0 &         127 &       75 &        0 &         0 \\
     &   & 4 & 20 &       3 &   492 &    0 &         446 &       45 &        0 &         0 \\
     &   &   & 40 &       0 &  3600 &   44 &        2299 &      180 &     1203 &         6 \\
\texttt{Wine} & 3 & 2 & 20 &       3 &   307 &    0 &           8 &       77 &      222 &       425 \\
     &   &   & 40 &       0 &  3600 &   10 &          88 &      180 &     3377 &        82 \\
     &   & 4 & 20 &       0 &  3600 &   12 &         288 &      180 &     3141 &        49 \\
     &   &   & 40 &       0 &  3600 &   76 &        1576 &      180 &     1925 &        10 \\
     & 5 & 2 & 20 &       2 &  1711 &    2 &          22 &      180 &     1502 &       784 \\
     &   &   & 40 &       0 &  3600 &   36 &         222 &      180 &     3262 &        35 \\
     &   & 4 & 20 &       0 &  3600 &   36 &         826 &      180 &     2644 &        46 \\
     &   &   & 40 &       0 &  3600 &   89 &        2792 &      180 &      722 &         3 \\
\bottomrule
\end{tabular}
\end{table}
\revision{In contrast to~\bnn{}, pre-processing time for~\enn{} consumes a considerably larger fraction of the total execution time across all instance configurations, particularly for the harder ones. The easier instances are almost exclusively solved in Phase 1 while the harder instances are often solved in Phase 2. For the \texttt{Wine} instances, the amount of nodes explored within the time limit in the branch-and-bound tree of Phase 2 decreases for larger problems, which indicates that it takes considerably more time to explore a node. This result is expected as problem \eqref{lagRelax}, solved at each node, becomes more computationally challenging as the number of networks, layers, or nodes increases. As mentioned above, the optimality gap for instances not solved within the time limit ranges on average between $2\%$ and $89\%$, which is considerably better than the gaps for \bnn{}.}

\subsection{Evaluation of the Acceleration Strategies} \label{sec:Sensitivity}
\revision{We conduct additional computational experiments to understand the effects of our different model enhancements on the performance of the algorithm.} 

\subsubsection{Experiments on Single Neural Networks}
\revision{Since both the targeted bound procedure and the valid inequalities also work for single NNs, we compare them against other methods from the literature used for single NNs. For each problem class, we generate three random test instances with~$\nnns = 1$, $\nlayers \in \{2,4\}$, and~$\nnodes{}= 50$, for a total of $36$ additional instances.}

\revision{In our first set of experiments, we compare our targeted bound approach, denoted as \texttt{TBP}, with two other approaches. The first one computes the values of $\LB$ and $\UB$ by solving 2 LPs per neuron and is denoted by \texttt{LP}. The second approach computes $\LB$ and $\UB$ by solving 2 MILPs per neuron, with a time limit of $5$ seconds per MILP, and is denoted by \texttt{MILP}. We design the following experiment to measure the performance of each method. We feed the values of~$\LB$ and~$\UB$ obtained by each of the methods to~\bnn{} and record the value of the bound obtained at the root node of the branch-and-bound tree. We also record the updated bound after adding $25,000$ cuts  
of the type proposed by \cite{anderson2020}. We then compute the percent improvement of each of the bounds obtained with respect to the (weaker) bounds delivered by~\texttt{LP} as follows:}
\begin{equation}
    \text{Percent bound improvement} = \frac{|\text{Bound} - \text{Bound } \texttt{LP} |}{|\text{Bound } \texttt{LP}|}
\end{equation}

\revision{Table \ref{tab:TBP} presents the results for this experiment aggregated  by problem class. Columns ``Root" report the improvement over the root node bound obtained by \texttt{LP} and columns ``Cuts" report the improvement to the bound after adding cuts. We also report the average time it takes to compute values for $\LB$ and $\UB$ in column ``Time", and the average number of MILPs solved in the pre-processing stage.   }
 \begin{table}[ht]
\centering
\scriptsize
\caption{Computational Time Break Down for the Enhanced algorithm.}
\label{tab:TBP}
\begin{tabular}{r|rrr|rrrr|rrrr}
\hline
&    \multicolumn{ 3}{c}{{\bf \texttt{LP}}} &               \multicolumn{ 4}{c}{{\bf \texttt{MILP}}} &                \multicolumn{ 4}{c}{{\bf \texttt{TBP}}} \\

Instance & Time &  Root & Cuts & Time &  Root & Cuts & MILPs &  Time &  Root &  Cuts &  MILPs \\
\hline
     \texttt{Beale} &          4 &        0\% &       78\% &         28 &       99\% &      100\% &        151 &         29 &       99\% &      100\% &         30 \\

     \texttt{Peaks} &          6 &        0\% &       66\% &        108 &       80\% &      100\% &        151 &        103 &       78\% &       96\% &         54 \\

      \texttt{Perm} &          7 &        0\% &       43\% &        196 &       68\% &       88\% &        151 &        139 &       66\% &       80\% &         58 \\

    \texttt{Spring} &          6 &        0\% &       29\% &        453 &       74\% &       89\% &        151 &        241 &       68\% &       80\% &         62 \\

  \texttt{Concrete} &          6 &        0\% &       17\% &        495 &       72\% &       72\% &        151 &        256 &       67\% &       67\% &         70 \\

      \texttt{Wine} &          8 &        0\% &       23\% &        556 &       74\% &       74\% &        151 &        292 &       68\% &       68\% &         77 \\
\hline
Average &          6 &       0\% &       43\% &        306 &       78\% &       87\% &        151 &        177 &       74\% &       82\% &         59 \\
\hline
\end{tabular}
\end{table}
\revision{The time used by \texttt{LP} is a fraction of the time used by \texttt{MILP} and \texttt{TBP}, as expected. The improvement to the bounds achieved by the cuts for \texttt{LP} is on average $43\%$. On the other hand, \texttt{MILP} exhibits the largest computational times and achieves an average bound improvement at the root node of $78\%$ before adding cuts, and of $87\%$ after adding the cuts, when compared to the bounds resulting from \texttt{LP}. Our approach achieves similar improvements of $74\%$ at the root node and $82\%$ after adding the cuts. Moreover, \texttt{TBP} uses considerably less computational time on average and solves fewer MILPs than \texttt{MILP}. We remark that depending on the definition of the critical neurons, controlled by parameter $\tau$, \texttt{TBP} could mimic the behaviour of \texttt{LP} for large values of $\tau$ and the behaviour of \texttt{MILP} for small values of $\tau$. We also note that the effect of the cuts is much more pronounced for \texttt{LP}, where the bounds after adding cuts are $43\%$ better, whereas for the other methods the additional improvement obtained from the cuts is only $9\%$ for \texttt{MILP} and $8\%$ for \texttt{TBP}. }    

\revision{In the second set of experiments, we  isolate the effect of the valid inequalities by comparing two approaches. We first compute the values of $\LB$ and $\UB$ by solving 2 LPs per neuron. Then, the first approach uses Formulation~\eqref{BigM} without any valid inequalities or any other enhancements; we refer to this algorithm as~\texttt{BigM}. The second approach includes valid inequalities~\eqref{VI} and is denoted as \texttt{BigM-VI}. We deactivate any preprocessing procedures and cuts from the solver for this experiment.} 
\revision{
We find that adding valid inequalities \eqref{VI} when bounds are computed via LPs leads to a moderate increase in the average computational time while obtaining virtually the same optimality gap across all problem classes on single NNs. This suggest that valid inequalities \eqref{VI} are too loose to have a strong effect on the computational performance of the algorithm and that strategies to strengthen their coefficients are needed to make them more impactful. A table with the summary of these experiments  is presented in the Appendix~\ref{app:Results}. }

\subsubsection{Analyzing the Lagrangian Relaxation-Based Approach}
\revision{We select a subset of $10$ instances that are solved to optimality within the time limit during Phase Two and analyze some additional performance measures to better understand the behaviour of our proposed approach (3 instances for Peaks, 5 instances for Perm, and 2 instances for Concrete). Table \ref{t17} reports the results for this experiment. The first column reports the average total computational time. The second column reports the optimality gap obtained at the end of Phase One. The third column presents the gap obtained at the beginning of Phase Two, after running the subgradient algorithm. The fourth column reports the average number of times that the algorithm stops branching and reverts to the big-M formulation (which happens when the domain of the variables becomes too small). Column five shows the average depth of the nodes explored in the branch-and-bound tree of Phase Two, and the last column reports the average number of nodes explored.}
\begin{table}[H]
\centering
\footnotesize
\caption{Analyzing the second phase of the algorithm.}
\label{t17}
\begin{tabular}{rrrrrrr}
\hline
 Instance & Time & Gap 1 & Gap 2  & \# BigM &  Depth & Nodes \\
\hline
\texttt{Peaks} &       1356 &       64\% &        1\% &                4 &          5 &         25 \\

\texttt{Perm} &       1198 &        9\% &       13\% &              47 &         12 &       1378 \\

\texttt{Concrete} &       1560 &       48\% &       20\% &               0 &         16 &        352 \\
\hline
Average &       1318 &       33\% &       11\% &             25 &         11 &        767 \\
\hline
\end{tabular}  
\end{table}
\revision{Table \ref{t17} shows that, on average, the bounds obtained after switching to Phase Two are considerably better than the bounds at the end of Phase One. This shows that the Lagrangian reformulation is on average tighter than the original formulation. There is an exception for instances of the \texttt{Perm} problem, in which the bounds at the end of Phase One are slightly better. This can be explained by the fact that, after terminating the MILP execution at the end of Phase One, we lose all the cuts and branching done by the solver up to that point.}

\revision{We also find that
the our algorithm reverts to the big-M formulation 
only for a relatively small fraction of the total number of nodes explored. In particular, this never happens for the~\texttt{Concrete} instances, meaning that all the nodes are pruned by bounds before ever reverting to the big-M formulation. We also observe that the average depth of the nodes explored is at most 16, which is a direct consequence of following the best-first node selection strategy. Finally, the average time used by the subgradient algorithm is negligible (i.e., less than one second) for all the problem instances thanks to our strategy of fixing the binary variables and solving LPs instead of MILPs. On the contrary, when solving MILPs without fixing any variables, the computational times for the subgradient algorithm increases to several minutes in some cases.     }

\section{Conclusion}\label{sec:conclusion}
Optimization models with embedded trained neural networks have been the focus of multiple studies from the literature in the past few years. We propose to use an ensemble of neural networks 
\revision{for optimization}. \revision{
Our experiments show that optimizing over ensembles delivers better and more robust results than optimizing over single NNs.}

\revision{We propose a Two Phase algorithm to solve optimization problems with objective functions represented by ensembles of neural networks}. 
We develop valid inequalities derived from a Benders decomposition approach and combine them with a targeted bound tightening procedure to reduce the computational times for the challenging optimization model at hand; \revision{these two techniques can also be applied to single NNs. Moreover, we explore the ensemble structure to develop a Lagrangian relaxation-based branch and bound algorithm, which is shown to improve considerably over a direct formulation.} Computational results show that our solution methods outperform an state-of-the-art branch-and-cut algorithm both in terms of CPU time and optimality gap, especially for large-sized ensembles of neural networks.

Our work opens different streams for future research. One direction relates to the interplay between the statistical properties of a trained neural network ensemble and its ensuing optimization model. We would like to investigate if predictors with lower MSE values directly result in optimization models with higher solution quality or which statistical performance measures of the predictors play a major role in ensuring that the optimization model produces high-quality solutions. 
Finally, we would like to explore alternative representations of the solution space which could result in tighter formulations (e.g., using decision diagrams to represent the space of feasible binary assignments of the auxiliary $z$-variables).

\bibliographystyle{plainnat}
\bibliography{nnEnsemble}

\newpage

\appendix

 \section{Detailed results} \label{app:Results}

\revision{
We show the complete results of our numerical experiments in
Tables~\ref{tab:BCresultsC} -- $\ref{t3}$. All entries associated with time are reported in seconds, and the optimality gaps are the percentage values reported by Gurobi. Average values are rounded to the closest integer value. Table~\ref{tab:BCresultsC} extends Table~\ref{tab:BCresultsBreakDown} (with all the results of~\bnn{}), and  Table~\ref{tab:2phaseresultsC} extends Table~\ref{tab:2phaseresultsBreakDown} (for~\enn{}); the columns in both tables are similar to their counterparts in the main text. Finally,  Table~\ref{t3} reports the results for the sensitivity analysis on the effect of the valid inequalities; in addition to the number of instances solved, the average solution time, and the average gap for each algorithm, we report the average number of cuts generated by~\texttt{BigM-VI}. }

 \begin{table}[H]
\centering
\scriptsize
\caption{Results for Baseline algorithm.}
\label{tab:BCresultsC}
\begin{tabular}{llll|rrrrr}
\toprule
     &   &   &    &  Solved &  Time &    Gap &  Preprocess &  B\&B Time \\
Instance &  $\nnns$ & $\nlayers$ & Nodes &         &       &        &             &           \\
\midrule
\texttt{Beale} & 3 & 2 & 20 &       3 &     1 &      0 &           0 &         0 \\
     &   &   & 40 &       3 &     8 &      0 &           1 &         8 \\
     &   & 4 & 20 &       3 &    29 &      0 &           1 &        27 \\
     &   &   & 40 &       0 &  3600 &   4242 &           7 &      3593 \\
     & 5 & 2 & 20 &       3 &     5 &      0 &           0 &         5 \\
     &   &   & 40 &       3 &    27 &      0 &           2 &        25 \\
     &   & 4 & 20 &       3 &   178 &      0 &           3 &       175 \\
     &   &   & 40 &       0 &  3600 &  27563 &          19 &      3581 \\
\texttt{Concrete} & 3 & 2 & 20 &       3 &    39 &      0 &           0 &        39 \\
     &   &   & 40 &       0 &  3600 &    166 &           2 &      3598 \\
     &   & 4 & 20 &       0 &  3600 &    237 &           3 &      3597 \\
     &   &   & 40 &       0 &  3600 &   1808 &          24 &      3576 \\
     & 5 & 2 & 20 &       0 &  3600 &     59 &           1 &      3599 \\
     &   &   & 40 &       0 &  3600 &    777 &           5 &      3595 \\
     &   & 4 & 20 &       0 &  3600 &    800 &           8 &      3592 \\
     &   &   & 40 &       0 &  3600 &   1355 &          66 &      3534 \\
\texttt{Peaks} & 3 & 2 & 20 &       3 &     3 &      0 &           0 &         3 \\
     &   &   & 40 &       3 &    42 &      0 &           1 &        40 \\
     &   & 4 & 20 &       3 &   359 &      0 &           2 &       357 \\
     &   &   & 40 &       0 &  3600 &    269 &          15 &      3585 \\
     & 5 & 2 & 20 &       3 &    13 &      0 &           1 &        12 \\
     &   &   & 40 &       3 &   184 &      0 &           4 &       180 \\
     &   & 4 & 20 &       0 &  3600 &    184 &           6 &      3595 \\
     &   &   & 40 &       0 &  3600 &    403 &          37 &      3563 \\
\texttt{Perm} & 3 & 2 & 20 &       3 &    10 &      0 &           0 &        10 \\
     &   &   & 40 &       2 &  2037 &     41 &           2 &      2035 \\
     &   & 4 & 20 &       1 &  3107 &     16 &           2 &      3105 \\
     &   &   & 40 &       0 &  3600 &    284 &          15 &      3585 \\
     & 5 & 2 & 20 &       3 &    62 &      0 &           1 &        61 \\
     &   &   & 40 &       0 &  3600 &     68 &           4 &      3596 \\
     &   & 4 & 20 &       0 &  3600 &     81 &           5 &      3595 \\
     &   &   & 40 &       0 &  3600 &    289 &          41 &      3559 \\
\texttt{Spring} & 3 & 2 & 20 &       3 &     2 &      0 &           0 &         2 \\
     &   &   & 40 &       1 &  2817 &     20 &           2 &      2816 \\
     &   & 4 & 20 &       1 &  2595 &    106 &           2 &      2592 \\
     &   &   & 40 &       0 &  3600 &   1508 &          20 &      3580 \\
     & 5 & 2 & 20 &       3 &    40 &      0 &           1 &        39 \\
     &   &   & 40 &       0 &  3600 &    116 &           4 &      3596 \\
     &   & 4 & 20 &       0 &  3600 &    282 &           7 &      3593 \\
     &   &   & 40 &       0 &  3600 &   6284 &          53 &      3547 \\
\texttt{Wine} & 3 & 2 & 20 &       3 &    78 &      0 &           0 &        78 \\
     &   &   & 40 &       0 &  3600 &    247 &           2 &      3598 \\
     &   & 4 & 20 &       0 &  3600 &    266 &           3 &      3597 \\
     &   &   & 40 &       0 &  3600 &  52442 &          25 &      3575 \\
     & 5 & 2 & 20 &       0 &  3600 &    103 &           1 &      3599 \\
     &   &   & 40 &       0 &  3600 &   2382 &           6 &      3594 \\
     &   & 4 & 20 &       0 &  3600 &  21991 &           8 &      3592 \\
     &   &   & 40 &       0 &  3600 &  21320 &          45 &      3555 \\
\bottomrule
\end{tabular}
\end{table}

 \begin{table}[H]
\centering
\scriptsize
\caption{Results for Enhanced algorithm.}
\label{tab:2phaseresultsC}
\begin{tabular}{llll|rrrrrrr}
\toprule
     &   &   &    &  Solved &  Time &  Gap &  Preprocess &  Phase 1 &  Phase 2 &  BB Nodes \\
Instance & $\nnns$ & $\nlayers$ & Nodes &         &       &      &             &          &          &           \\
\midrule
\texttt{Beale} & 3 & 2 & 20 &       3 &     1 &    0 &           1 &        0 &        0 &         0 \\
     &   &   & 40 &       3 &     7 &    0 &           6 &        1 &        0 &         0 \\
     &   & 4 & 20 &       3 &    16 &    0 &          12 &        5 &        0 &         0 \\
     &   &   & 40 &       3 &   144 &    0 &         106 &       38 &        0 &         0 \\
     & 5 & 2 & 20 &       3 &     5 &    0 &           4 &        0 &        0 &         0 \\
     &   &   & 40 &       3 &    20 &    0 &          16 &        4 &        0 &         0 \\
     &   & 4 & 20 &       3 &    29 &    0 &          24 &        4 &        0 &         0 \\
     &   &   & 40 &       3 &   375 &    0 &         312 &       64 &        0 &         0 \\
\texttt{Concrete} & 3 & 2 & 20 &       3 &    18 &    0 &           6 &       12 &        0 &         0 \\
     &   &   & 40 &       2 &  2117 &    2 &          67 &      180 &     1865 &       367 \\
     &   & 4 & 20 &       0 &  3600 &    5 &         291 &      180 &     3118 &       220 \\
     &   &   & 40 &       0 &  3600 &   71 &        1477 &      180 &     1985 &        10 \\
     & 5 & 2 & 20 &       1 &  2971 &    1 &          16 &      180 &     2762 &     12787 \\
     &   &   & 40 &       1 &  3040 &    1 &         148 &      180 &     2698 &       334 \\
     &   & 4 & 20 &       0 &  3600 &   16 &         780 &      180 &     2632 &       125 \\
     &   &   & 40 &       0 &  3600 &   79 &        2681 &      180 &      842 &         4 \\
\texttt{Peaks} & 3 & 2 & 20 &       3 &     4 &    0 &           3 &        0 &        0 &         0 \\
     &   &   & 40 &       3 &    22 &    0 &          16 &        6 &        0 &         0 \\
     &   & 4 & 20 &       3 &    40 &    0 &          35 &        5 &        0 &         0 \\
     &   &   & 40 &       3 &   824 &    0 &         505 &      180 &      139 &        19 \\
     & 5 & 2 & 20 &       3 &     7 &    0 &           7 &        1 &        0 &         0 \\
     &   &   & 40 &       3 &    60 &    0 &          44 &       16 &        0 &         0 \\
     &   & 4 & 20 &       3 &   129 &    0 &         106 &       23 &        0 &         0 \\
     &   &   & 40 &       3 &  1381 &    0 &         979 &      180 &      221 &        25 \\
\texttt{Perm} & 3 & 2 & 20 &       3 &     7 &    0 &           3 &        4 &        0 &         0 \\
     &   &   & 40 &       3 &    64 &    0 &          21 &       43 &        0 &         0 \\
     &   & 4 & 20 &       3 &    89 &    0 &          37 &       52 &        0 &         0 \\
     &   &   & 40 &       1 &  3189 &    1 &         832 &      180 &     2192 &       150 \\
     & 5 & 2 & 20 &       3 &    14 &    0 &           7 &        6 &        0 &         0 \\
     &   &   & 40 &       3 &   966 &    0 &          70 &      180 &      716 &      1863 \\
     &   & 4 & 20 &       3 &   457 &    0 &         110 &      150 &      196 &       357 \\
     &   &   & 40 &       0 &  3600 &    4 &        1608 &      180 &     1795 &        66 \\
\texttt{Spring} & 3 & 2 & 20 &       3 &     7 &    0 &           6 &        1 &        0 &         0 \\
     &   &   & 40 &       3 &    69 &    0 &          44 &       25 &        0 &         0 \\
     &   & 4 & 20 &       3 &   172 &    0 &         144 &       28 &        0 &         0 \\
     &   &   & 40 &       0 &  3600 &   41 &        1198 &      180 &     2284 &        12 \\
     & 5 & 2 & 20 &       3 &    18 &    0 &          13 &        5 &        0 &         0 \\
     &   &   & 40 &       3 &   202 &    0 &         127 &       75 &        0 &         0 \\
     &   & 4 & 20 &       3 &   492 &    0 &         446 &       45 &        0 &         0 \\
     &   &   & 40 &       0 &  3600 &   44 &        2299 &      180 &     1203 &         6 \\
\texttt{Wine} & 3 & 2 & 20 &       3 &   307 &    0 &           8 &       77 &      222 &       425 \\
     &   &   & 40 &       0 &  3600 &   10 &          88 &      180 &     3377 &        82 \\
     &   & 4 & 20 &       0 &  3600 &   12 &         288 &      180 &     3141 &        49 \\
     &   &   & 40 &       0 &  3600 &   76 &        1576 &      180 &     1925 &        10 \\
     & 5 & 2 & 20 &       2 &  1711 &    2 &          22 &      180 &     1502 &       784 \\
     &   &   & 40 &       0 &  3600 &   36 &         222 &      180 &     3262 &        35 \\
     &   & 4 & 20 &       0 &  3600 &   36 &         826 &      180 &     2644 &        46 \\
     &   &   & 40 &       0 &  3600 &   89 &        2792 &      180 &      722 &         3 \\
\bottomrule
\end{tabular}
\end{table}

\begin{table}[H]
\centering
\footnotesize
\caption{Assessing the effect of the valid inequalities.}
\label{t3}
\begin{tabular}{r|rrr|rrrr}
\hline
           &      \multicolumn{ 3}{c}{{\bf \texttt{BigM}}} &                \multicolumn{ 4}{c}{{\bf \texttt{BigM-VI}}} \\






Instance & Solved & Time &  Gap & Solved &  Time &   Gap &  \# Cuts \\
\hline
     \texttt{Beale} &         6 &          17 &        0\% &         6&      17 &                 0\% &          5 \\

     \texttt{Peaks} &       3 &           1868 &       85\% &       3&       1858 &                86\% &         13 \\

      \texttt{Perm} &        6 &          453 &        0\% &        6&       804 &                  0\% &          7 \\

    \texttt{Spring} &    3   & 1803 &                 21\% &       3&        1803 &                21\% &          9 \\

  \texttt{Concrete} &     3  & 1802 &                 39\% &       3&        1802 &                38\% &          7 \\

      \texttt{Wine} &      3 & 2379 &                 96\% &       3&        2465 &                96\% &         15 \\
\hline
   Average &      24 & 1387 &                40\% &       24&       1458 &                40\% &          9 \\
\hline
\end{tabular}  

\end{table}

\section{Details of the Hyper Parameters Configuration} \label{app: hyperparameter}
\revision{
Table~\ref{tab:hyperparameter} reports the hyper parameters of the selected configurations (see Section~\ref{subsec:solqualityCompare}). The~\texttt{LHS} columns contain NNs trained on latin hypercube sampled dataset, whereas the~\texttt{MVN} columns contain NNs trained on data set sampled from multivariate normal distribution. Observe that~$e=1$ represents single NNs, whereas $e > 1$ represents ensembles.}
 \begin{table}[H]
\centering
\scriptsize
\caption{Hyper parameters of selected configurations}
\label{tab:hyperparameter}
\begin{tabular}{c|c|c|c|c|c| c|c|c|c|c}
\hline
&  \multicolumn{ 5}{c|}{\texttt{LHS}} & \multicolumn{5}{c}{\texttt{MVN}} \\
\hline
Instance & $\nnns$ & $\nlayers$ & Nodes &  Learning Rate  & Batch Size 
         & $\nnns$ & $\nlayers$ & Nodes &  Learning Rate  & Batch Size\\
\hline
\texttt{Peaks} & 2 & 2 & 145 &   0.0008035   &    32
             & 2 & 2 & 145 &   0.0011023   &    32 \\
             
               & 3 & 2 & 105 &   0.0009381   &    32
               & 3 & 2 & 115 &   0.0009605   &    32 \\
               
               & 3 & 2 & 90  &   0.0016427  &    32
               & 2 & 2 & 105  &   0.0011276  &    32 \\
               
               & 3 & 2 & 100 &   0.0007235   &    32
               & 2 & 2 & 130 &   0.0009183   &    32 \\
               
               & 1 & 2 & 140 &   0.0020914   &    32
               & 1 & 2 & 130 &   0.0021041   &    32 \\
               
               & 1 & 2 & 150 &   0.0005355   &    32
               & 1 & 2 & 115 &   0.0011318   &    32 \\
               
               & 1 & 2 & 130  &  0.0014449  &    32 
               & 1 & 2 & 105 &  0.0015904  &    32 \\
               
               & 1 & 2 & 125 &   0.0007010   &    32
               & 1 & 2 & 125 &   0.0009507   &    32 \\
               \midrule
\texttt{Beale} & 4 & 2 & 85 &    0.0008707   &    32 
               & 2 & 2 & 120 &   0.0016596   &    32 \\
               & 2 & 2 & 65 &    0.0010586   &    32 
               & 2 & 2 & 140 &   0.0019661   &    32 \\
               & 3 & 2 & 60  &   0.0021587  &    64 
               & 2 & 2 & 85  &   0.0017663  &    32 \\
               
               & 4 & 2 & 75 &    0.0022218   &    32 
               & 2 & 2 & 75 &   0.0019195   &    32 \\
               
               & 1 & 2 & 105 &   0.0020312   &    64 
               & 1 & 2 & 140 &   0.0026932   &    32 \\
               
               & 1 & 2 & 85 &    0.0004390   &    32
               & 1 & 2 & 80 &   0.0037704   &    32 \\
               
               & 1 & 2 & 125  &  0.0003647  &    64 
               & 1 & 2 & 90  &  0.0020293  &    32 \\
               
               & 1 & 2 & 95 &    0.0004334   &    64
               & 1 & 2 & 115 &   0.0025659   &    32 \\
               \midrule
\texttt{Perm}  & 3 & 2 & 65 &     0.0005243   &    32 
                & 2 & 2 & 90 &   0.0003264   &    64 \\

               & 3 & 2 & 100 &    0.0004778   &    64 
               & 2 & 2 & 115 &   0.0004152   &    32 \\
               
               & 4 & 2 & 45  &   0.0007861  &    32
               & 2 & 2 & 150  &   0.0003464  &    64 \\
               
               & 3 & 2 & 85 &    0.0007125   &    32
               & 3 & 2 & 80 &   0.0006148   &    64 \\
               
               & 1 & 2 & 70 &   0.0007341   &    32
               & 1 & 2 & 100 &   0.0001573   &    32 \\
               
               & 1 & 2 & 105 &    0.0005455   &    64
               & 1 & 2 & 150 &   0.0003580   &    32 \\
               
               & 1 & 2 & 120  &  0.0004419  &    64
               & 1 & 2 & 70  &  0.0003460  &    32 \\
               
               & 1 & 2 & 95 &    0.0005871   &    32
               & 1 & 2 & 90 &   0.0003411   &    32 \\
               \midrule
\texttt{Spring}& 3 & 2 & 175 &    0.0052601   &    32
                & 2 & 2 & 115 &   0.0021206   &    32 \\
                
               & 2 & 2 & 90 &    0.0016923   &    32 
               & 3 & 2 & 145 &   0.0015516   &    32 \\
               
               & 2 & 2 & 115  &   0.0021206  &    32
               & 3 & 2 & 110  &   0.0023950  &    32 \\
               
               &2 & 2 & 150 &    0.0019307   &    32
               & 3 & 2 & 180 &   0.0021299   &    32 \\
               
               &1 & 2 & 245 &   0.0032445   &    32
               & 1 & 2 & 140 &   0.0017628   &    32 \\
               
               &1 & 2 & 130 &    0.0016556   &    32
               & 1 & 2 & 105 &   0.0016161   &    32 \\
               
               &1 & 2 & 190  &  0.0035503  &    32
               & 1 & 2 & 215  &  0.0014401  &    32 \\
               
               &1 & 2 & 90 &    0.0022293   &    32
               & 1 & 2 & 175 &   0.0015953   &    32 \\
               \midrule
\end{tabular}
\end{table}

\end{document}